\documentclass[a4paper,fleqn]{cas-dc}

\usepackage[numbers]{natbib}
\usepackage{booktabs}
\usepackage{tabularx}
\usepackage{multirow}
\usepackage{xcolor}
\usepackage{array}
\usepackage{amsmath}
\usepackage{amssymb}
\usepackage{url}
\usepackage{graphicx}
\usepackage{placeins}
\usepackage{capt-of}


\begin{document}
\let\WriteBookmarks\relax
\def\floatpagepagefraction{1}
\def\textpagefraction{.001}

\shorttitle{Context fusion for endoscopic polyp reporting}
\shortauthors{Yang et~al.}

\title[mode=title]{From Generalist to Specialist: A Context-Fusion Framework for Endoscopic Polyp Reporting with a Frozen VLM}



\author[1,2,3,4]{Ruijie Yang}
\fnmark[1]

\author[5,6]{Yan Zhu}
\fnmark[1]

\author[5,6]{Peiyao Fu}

\author[3,4]{Siyuan Li}

\author[3,4]{Te Luo}

\author[1,2]{Zhihua Wang}

\author[5,6]{Quanlin Li}

\author[5,6]{Pinghong Zhou}
\cormark[1]
\ead{zhou.pinghong@zs-hospital.sh.cn}

\author[7]{Xian Yang}
\cormark[1]
\ead{xian.yang@manchester.ac.uk}

\author[3,4]{Shuo Wang}
\cormark[1]
\ead{shuowang@fudan.edu.cn}

\affiliation[1]{organization={Zhejiang University},
                city={Hangzhou},
                country={China}}

\affiliation[2]{organization={Shanghai Institute for Advanced Study, Zhejiang University},
                city={Shanghai},
                country={China}}

\affiliation[3]{organization={Shanghai Key Laboratory of MICCAI},
                city={Shanghai},
                country={China}}

\affiliation[4]{organization={Digital Medical Research Center, School of Basic Medical Sciences, Fudan University},
                city={Shanghai},
                country={China}}

\affiliation[5]{organization={Endoscopy Center and Endoscopy Research Institute, Zhongshan Hospital, Fudan University},
                city={Shanghai},
                country={China}}

\affiliation[6]{organization={Shanghai Collaborative Innovation Center of Endoscopy},
                city={Shanghai},
                country={China}}

\affiliation[7]{organization={Alliance Manchester Business School, The University of Manchester},
                city={Manchester},
                country={United Kingdom}}

\fntext[fn1]{These authors contributed equally to this work.}
\cortext[cor1]{Corresponding authors}

\begin{abstract}
Reliable endoscopic polyp reporting requires integrating quantitative lesion sizing, standardized Paris classification, and clinically meaningful morphological description within a single record. General-purpose vision-language models (VLMs) offer a unified interface for image understanding and report generation. Existing specialization strategies, however, typically rely on task-specific models or model-weight adaptation, leaving unresolved how to introduce reliable specialist knowledge while preserving both this unified interface and the VLM's pretrained capabilities. We introduce a context-fusion framework that specializes a frozen general purpose VLM through both implicit instruction context and explicit transduction context without modifying its pretrained weights. Specifically, a self-supervised polyp encoder retrieves related image--report pairs as explicit, query-specific evidence, while learned continuous specialist tokens provide implicit instruction context shared across cases. Experiments were conducted on 2,056 expert-annotated public endoscopic images. We compared the framework with general-purpose VLMs, task-specific predictors, and weight-adaptation methods to assess specialist performance, unified reporting, and adaptation efficiency. Across numerical, categorical, and report-generation metrics, the proposed framework substantially improved direct frozen-VLM inference and achieved the strongest overall performance among the evaluated methods. It added trainable parameters equal to only 0.006\% of the frozen VLM's parameter count. When the top-1 retrieved case carried the correct target category, our framework corrected 70.5\% of the errors made by a weight-adaptation baseline. These findings support the context-fusion framework as a lightweight and effective strategy for specialist adaptation of a frozen VLM.

\end{abstract}


\maketitle

\section{Introduction}
Colonoscopy plays a central role in colorectal-cancer prevention by enabling polyp detection and the removal of premalignant lesions \cite{zauber2012polypectomy}. Within this clinical workflow, reliable polyp reporting supports lesion management, post-polypectomy surveillance, clinical communication, and quality review \cite{coe2012reporting,usmstf2020}. Reporting standards require explicit documentation of polyp size and morphology \cite{esgeQuality2017}, while the Paris classification standardizes macroscopic lesion type \cite{parisClassification2005}. Automated polyp reporting is challenging because several clinically distinct judgments must be made from a single monocular image. The system must estimate lesion size without an intrinsic scale, distinguish subtle morphological patterns for Paris classification, and describe surface, base, border, and associated findings in clinically meaningful language \cite{antonelli2026polypSize,massimi2025parisMLLM}. 

Existing approaches to automated polyp reporting broadly follow two paradigms. Modular pipelines \cite{qu2021endoscopyReport} predict predefined clinical attributes and assemble them into a template-based report, making selected findings directly assessable but limiting the report to a fixed set of outputs. End-to-end image-to-text systems \cite{colorectalPolypBasicDescription,zhang2026samColonPolypGen} generate more flexible descriptions through a unified interface. However, errors in size or type can remain hidden within otherwise fluent reports \cite{miura2021factual}. Neither paradigm therefore combines unified multi-task reporting with direct verification of quantitative, categorical, and descriptive outputs.

Recent advances in general-purpose VLMs \cite{moor2023generalist,biomedgpt2024} offer a promising foundation for solving medical reporting tasks. Direct prompting generalist VLMs, however, does not reliably provide the specialist knowledge needed to interpret subtle endoscopic findings, and hallucinations remain a concern \cite{medvh2025}. On the other hand, conventional specialization adapts model weights to domain data, which may alter pretrained reasoning capacity and response diversity \cite{lou2026conceptvlm}. The unresolved challenge is therefore to introduce reliable specialist knowledge into a general-purpose VLM without modifying its pretrained weights, while preserving its pretrained capabilities and unified generative interface.

\begin{figure*}[!t]
\centering
\includegraphics[width=\textwidth]{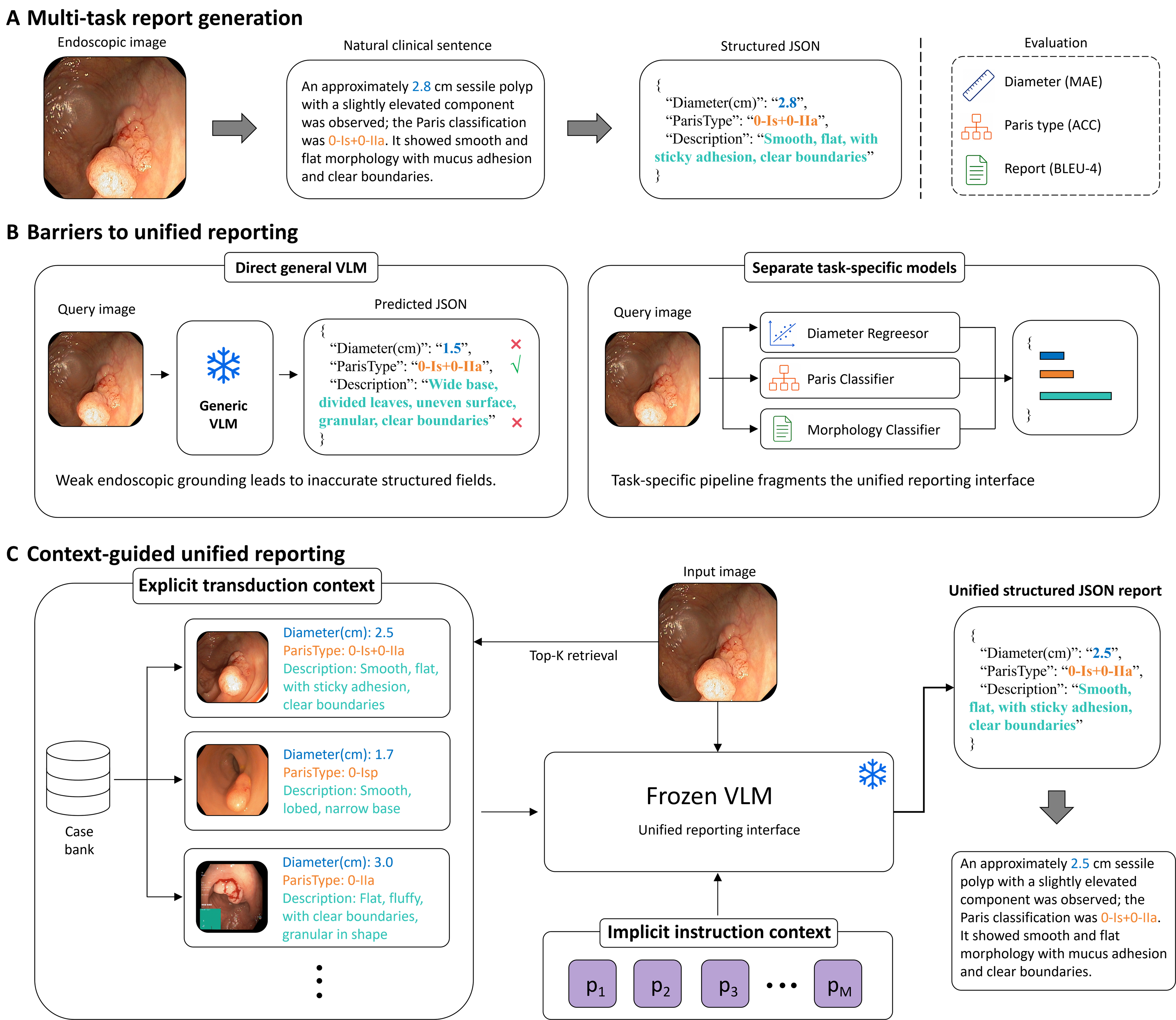}
\caption{Workflow and motivation for unified endoscopic polyp report generation. (A) Conventional image-to-report generation embeds quantitative, categorical, and descriptive findings within a natural clinical sentence. We reorganize the same clinical content as a structured multi-task JSON report: diameter and Paris type can be evaluated directly, while the complete output is converted into a fixed clinical sentence for report-generation evaluation. (B) Direct general-VLM prompting lacks specialist case context, whereas separate task-specific models fragment reporting across independent predictors. (C) The proposed framework combines explicit transduction context and implicit instruction context around a frozen VLM, enabling all three reporting tasks through one unified interface.}
\label{fig:workflow}
\end{figure*}

To address this challenge, we introduce a context-fusion framework that equips a frozen general-purpose VLM with specialist context rather than adapting its pretrained weights. The framework combines two forms of specialist context. \textbf{Explicit transduction context} consists of retrieved image--report pairs that supply query-specific clinical evidence. \textbf{Implicit instruction context} is represented by learned continuous specialist tokens that provide persistent reporting guidance across cases. At inference, the query image and both forms of specialist context are jointly presented to the frozen VLM under a structured instruction, which prompts it to estimate diameter, classify Paris type, and generate a morphological description in a single JSON report. The framework thereby supports multiple specialized reporting tasks through one generative interface. Because specialization is confined to the context, it requires no report-specific changes to the VLM architecture and leaves all pretrained VLM weights unchanged. Figure~\ref{fig:workflow} summarizes the reporting task, the limitations of the two existing paradigms, and the proposed context-based alternative.

We instantiate this hypothesis using latest open-source VLM \cite{qwen35_2026} and 2,056 public endoscopic images with in-house expert annotations. We compare the proposed framework with directly prompted general-purpose VLMs, weight-adaptation methods and modular designs with task-specific predictors. Across the validation and test sets, the framework achieves the strongest overall performance among the evaluated methods while adding only 0.5~M trainable parameters. Ablation studies further show that the two contexts provide complementary gains and that retrieval benefits depend on evidence relevance rather than input length.

Our contributions are threefold:
\begin{itemize}
    \item We formulate multi-task endoscopic polyp reporting as unified generation of continuous diameter, categorical Paris type, and open-ended morphological description within one structured schema. We further construct and release a 2,056-image benchmark with expert annotations.

    \item We introduce a context-fusion framework that combines explicit transduction context with implicit instruction context. The framework confines specialist adaptation to the input context and leaves all pretrained VLM weights unchanged.

    \item We systematically compare different strategies adapting general-purpose VLMs. The proposed context-fusion framework achieves the best overall performance while maintaining high efficiency.
\end{itemize}

\section{Related work}

\subsection{Automated endoscopic and polyp reporting}

Prior work on automated endoscopic reporting broadly follows two design patterns: assembling reports from task-specific predictions or generating report text directly from images. Qu \emph{et al.} \cite{qu2021endoscopyReport} combined multiple recognition models with rule-based report assembly and directly evaluated lesion classification and size error, but the generated reports were confined to predefined recognizers and report fields. In contrast, Fonoll\`a \emph{et al.} \cite{colorectalPolypBasicDescription} used an image encoder and a BERT language module to generate BASIC-based polyp descriptions. As automated report generation has advanced, rigorous assessment of report quality has emerged as a distinct methodological challenge. Metrics such as BLEU, ROUGE-L, and METEOR were used to measure resemblance to reference text, but did not separately expose errors in quantitative measurements or categorical findings. SAM-ColonPolypGen \cite{zhang2026samColonPolypGen} supplemented language metrics with two factual checks: a generated size was correct when it differed from the reference by no more than 0.1~cm, and morphology required an exact label match. Because these values were embedded in free text, the metric definitions assumed that a single numeric value and morphology label could first be recovered from the report. Omissions, malformed values, or ambiguous mentions would still require an additional matching policy. Report-Angel \cite{reportAngel2026} turned to expert subjective assessment of accuracy and completeness. Experts judged whether a report was clinically acceptable and rated sentence coherence, feature completeness, and feature accuracy, while quantitative lesion sizing or categorical morphology classification were not evaluated. Existing systems therefore either make selected findings directly verifiable within a restricted schema or generate richer prose whose key facts require additional extraction. A shared public benchmark that jointly evaluates regression, standardized classification, and narrative description remains absent.

\subsection{Medical specialization of VLMs}

Adapting general-purpose VLMs to the medical domain has generally relied on large-scale, domain-specific multimodal data, particularly for continued pretraining and multimodal instruction tuning. Medical VLMs such as LLaVA-Med \cite{llavamed2023}, BiomedGPT \cite{biomedgpt2024}, and Hulu-Med \cite{hulumed2025} acquire broad medical capability from large, heterogeneous image--text corpora. When the target is a narrower reporting domain, adaptation is usually restricted to part of the model. Flamingo-CXR \cite{tanno2025flamingoCXR} trains its visual and connective components while freezing the language model, whereas PeFoMed \cite{pefomed2026} learns a visual projection and applies LoRA without updating the vision or language backbones. Prompt tuning \cite{lester2021promptTuning} and visual prompt tuning \cite{jia2022vpt} reduce the trainable component further to continuous tokens. These parameter-efficient strategies lower optimization cost, but they do not remove the need for representative specialist supervision. The effect of limited data is evident in prior studies: Flamingo-CXR \cite{tanno2025flamingoCXR} observed overfitting when its language component was updated, and ConceptVLM \cite{lou2026conceptvlm} found that direct specialist fine-tuning reduced the general analytical capability of the underlying VLM. This constraint is particularly relevant to endoscopy, where expert image--report pairs are less abundant than in radiology. Report-Angel \cite{reportAngel2026}, one of the few endoscopy-specific reporting systems, was trained on 20,617 image--text pairs and combined a fine-tuned MLLM with conventional models; meanwhile, GI-Bench \cite{zhu2026gibench} identifies substantial gaps in visual grounding and factual correctness among general-purpose models. Reliable specialist adaptation from smaller endoscopic reporting datasets therefore remains insufficiently studied.

\subsection{Context engineering for large models}

Context engineering \cite{mei2025contextEngineering} organizes instructions, demonstrations, retrieved evidence, and learned context around a pretrained model. Existing methods instantiate this principle at different levels, including instruction optimization, demonstration selection, and external knowledge retrieval. At the instruction level, Automatic Prompt Engineer \cite{zhou2023ape} generates and selects candidate instructions, while OPRO \cite{yang2024opro} iteratively improves instructions using the scores of earlier candidates. Their final prompts provide instruction guidance shared across queries rather than query-specific evidence. At the demonstration level, LLM-R \cite{wang2024retrieveICL} trains a dense retriever from LLM feedback to select input--output examples, whereas visual in-context learning \cite{zhou2024visualICL} retrieves, summarizes, and composes visual demonstrations. Beyond instructions and demonstrations, retrieval can supply a model with external knowledge relevant to the current query. RAG \cite{lewis2020rag} retrieves external text passages, whereas GraphRAG \cite{edge2024graphRAG} organizes document collections into entity graphs and community summaries. BiomedKAI \cite{biomedKAI2026} extends text retrieval to biomedical knowledge graphs with query-aware retrieval and specialized agents; these systems retrieve textual knowledge rather than paired clinical images and reports. Medical multimodal RAG has moved closer to case-based evidence. CXR-RePaiR \cite{endo2021cxrRepair} retrieves reference reports for chest radiographs, MMed-RAG \cite{mmedrag2025} uses domain-aware retrieval and adaptive context selection on radiology, ophthalmology, and pathology datasets for medical VQA and report generation, and FactMM-RAG \cite{sun2025factmmrag} trains a fact-aware retriever for radiology report generation. In endoscopy, EndoFinder \cite{yang2024endofinder} performs image-to-image retrieval of visually similar colorectal polyps to support explainable diagnosis. Taken together, prior work has explored shared instruction guidance, query-specific demonstrations, and multimodal retrieval largely as separate forms of context. Their integration within a frozen VLM remains underexplored for endoscopic reporting.

\begin{figure*}[!t]
\centering
\includegraphics[width=\textwidth]{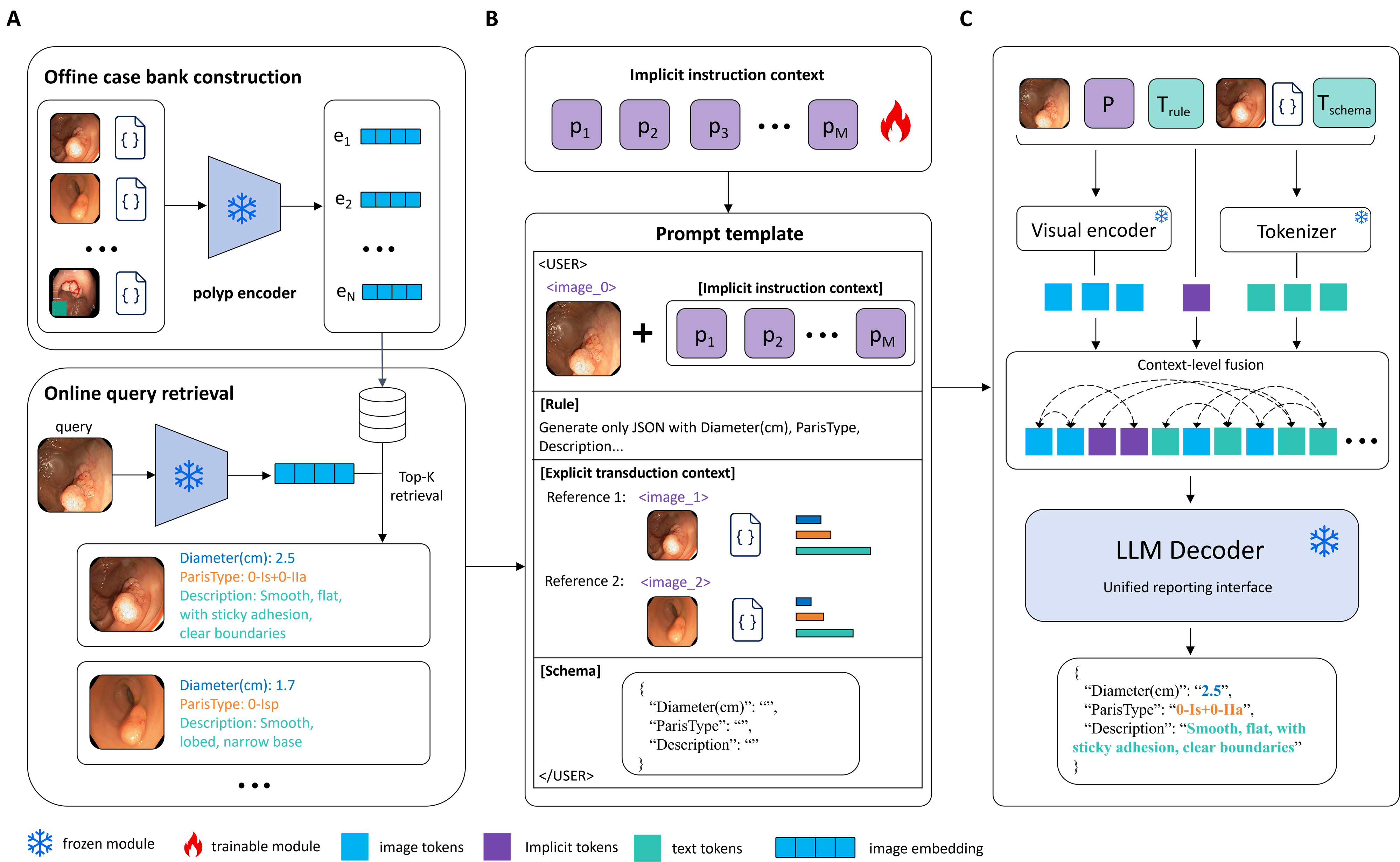}
\caption{Context-fusion framework for a frozen VLM. (A) Offline, a fixed self-supervised polyp encoder maps the training images to L2-normalized CLS representations $e_1,\ldots,e_N$, which are stored with their reports in the case bank; online, the same encoder retrieves the top-$K$ image--report pairs for a query. (B) The retrieved pairs provide explicit transduction context, while learned continuous specialist tokens provide implicit instruction context within the multimodal prompt. (C) The composed context is supplied to the frozen VLM to generate one structured report.}
\label{fig:method}
\end{figure*}

\section{Method}

\subsection{Problem formulation}

Let $\mathcal{D}=\{(x_i,y_i)\}_{i=1}^{n}$ denote the reporting dataset, where $x_i$ is an endoscopic image and $y_i=(d_i,p_i,r_i)$ is its structured report. The report contains three fields corresponding to three tasks: diameter estimation $d_i$, Paris-type classification $p_i$, and free-text morphology description $r_i$. Given a query image $x_q$, the objective is to generate $\hat{y}_q=(\hat{d}_q,\hat{p}_q,\hat{r}_q)$ in one autoregressive response. Let $G_{\theta}$ denote a pretrained VLM with parameters $\theta$. Conventional adaptation uses $\mathcal{D}$ to update $\theta$ for the target domain. In contrast, we keep $\theta$ fixed and formulate specialist reporting as context-conditioned generation:
\begin{equation}
\hat{y}_q=G_{\theta}(x_q,C_q),
\end{equation}
where $C_q$ denotes the additional context supplied with the query. Adaptation therefore changes the information available to $G_{\theta}$ rather than its pretrained parameters.

\subsection{Framework overview}

Our framework specializes frozen VLMs through two complementary forms of context (Fig.~\ref{fig:method}). The first is query-specific: explicit transduction context retrieves prior image--report pairs that demonstrate how visually related lesions were described and assessed. The second is shared across the dataset: implicit instruction context uses learned continuous specialist tokens to condition the frozen VLM toward the reporting problem. These sources are combined with the query image, textual reporting rules, and output schema in one multimodal input. The frozen VLM receives this composed context and generates all three report fields in a single response. The framework therefore separates specialist adaptation from the VLM weights: case-specific information is supplied through retrieval, whereas shared reporting information is represented by a compact trainable context.

\subsection{Explicit transduction context}

We first re-index the $N$ training image--report pairs as $\{(x_i,y_i)\}_{i=1}^{N}$. 
For each pair, a fixed image encoder $E(\cdot)$ produces an L2-normalized CLS representation $e_i=E(x_i)$. EndoFinder \cite{yang2024endofinder}, a self-supervised polyp encoder trained on polyp images is utilized in this study. We compute these representations once and construct the case bank $\mathcal{B}=\{(x_i,y_i,e_i)\}_{i=1}^{N}$. For a query image $x_q$, we obtain $e_q=E(x_q)$ and calculate its similarity to the $i$th bank image as
\begin{equation}
s_i=e_q^\top e_i,
\end{equation}
which is cosine similarity because both representations are normalized. We select $K$ image--report pairs and retain their scores as the query-specific retrieval set
\begin{equation}
\mathcal{R}_q=\big[(x_{r_j},y_{r_j},s_{r_j})\big]_{j=1}^{K}.
\end{equation}
During training, $K$ cases are randomly sampled from the ten highest-scoring candidates. This prevents the same evidence set from being used for a query in every epoch. At inference, $\mathcal{R}_q$ contains the $K$ highest-scoring cases in descending similarity order.

For each retrieved pair, let $\mathbf{v}_{r_j}$ denote the visual-token sequence of image $x_{r_j}$ and let $\mathbf{t}_{r_j}$ denote the text-token sequence obtained from report $y_{r_j}$. The pair and its similarity score form one multimodal evidence block:
\begin{equation}
B_j=\big[\mathbf{v}_{r_j},\mathbf{t}_{r_j},s_{r_j}\big], \qquad
\mathcal{E}_q=[B_1,\ldots,B_K].
\end{equation}
Each block retains the association between the retrieved image, its report, and its similarity score. In the implemented prompt, the retrieved cases are numbered by rank so that each image is explicitly paired with its report and score. A retrieved image supplies a visual analogue of the query, while its paired report provides the corresponding diameter, Paris type, and clinical description. The pair therefore demonstrates an observed mapping from endoscopic appearance to specialist report rather than supplying isolated labels or ungrounded text. We refer to these inspectable, query-dependent cases as explicit transduction context.

\subsection{Implicit instruction context}

We employ prompt tuning technique to parameterize the implicit instruction context. It is a sequence of continuous specialist tokens $P=[p_1,\ldots,p_M]\in\mathbb{R}^{M\times h}$, where $M$ is the prompt length and $h$ is the hidden dimension of the frozen VLM. The same $P$ is shared across all cases and is optimized from the report-generation objective while $G_{\theta}$ remains fixed. Because these tokens are learned from all training reports, they provide shared conditioning for producing the required numerical, categorical, and descriptive fields. Unlike a textual instruction, these continuous vectors are learned directly in the VLM embedding space; unlike retrieved cases, they contain no identifiable image, report, or query-specific fact.

The specialist tokens are inserted immediately after the query-image tokens. Under causal attention, their hidden states can depend on the current image before the later instruction and retrieval tokens are processed. Thus, although their learned embeddings are shared across cases, their contextualized representations are conditioned on the query. The parameterization follows prompt tuning, but its role in the framework is specifically to provide case-shared implicit context that complements the case-specific evidence supplied by retrieval. It is trained under the same multimodal evidence composition used for report generation rather than as an isolated textual prompt.

\subsection{Context composition and fusion}

Let $\mathbf{v}_q$ denote the visual-token sequence of query image $x_q$, and let $T_{\mathrm{rule}}$ and $T_{\mathrm{schema}}$ denote the token sequences of the textual reporting rules and output schema, respectively. The context for each query is composed as
\begin{equation}
C_q=[\mathbf{v}_q,P,T_{\mathrm{rule}},\mathcal{E}_q,T_{\mathrm{schema}}].
\end{equation}
The query-image tokens $\mathbf{v}_q$ appear first because they represent the case to be reported. The specialist tokens $P$ follow immediately, allowing their hidden states to be conditioned on the query image. The textual-rule tokens $T_{\mathrm{rule}}$ then define the three reporting tasks before the retrieved evidence $\mathcal{E}_q$ is introduced. Each evidence block $B_j\in\mathcal{E}_q$ contributes one retrieved image, its paired report, and its similarity score. Finally, the output-schema tokens $T_{\mathrm{schema}}$ specify the required keys and close the user prompt before generation begins. In implementation, the multi-image processor serializes all image placeholders before the textual prompt body; rank labels preserve the one-to-one correspondence between every retrieved image and its report.

This ordering creates a single context in which each source has a distinct role. The query image supplies the primary observation; $P$ provides shared specialist conditioning; the retrieved pairs provide query-specific precedents; and the textual rules constrain the reporting task and response structure. We refer to their joint use within the same multimodal input as context fusion. During autoregressive generation, the frozen VLM conditions each report token on this composed context and produces one structured multi-task report.

\begin{figure*}[!t]
\centering
\includegraphics[width=\textwidth]{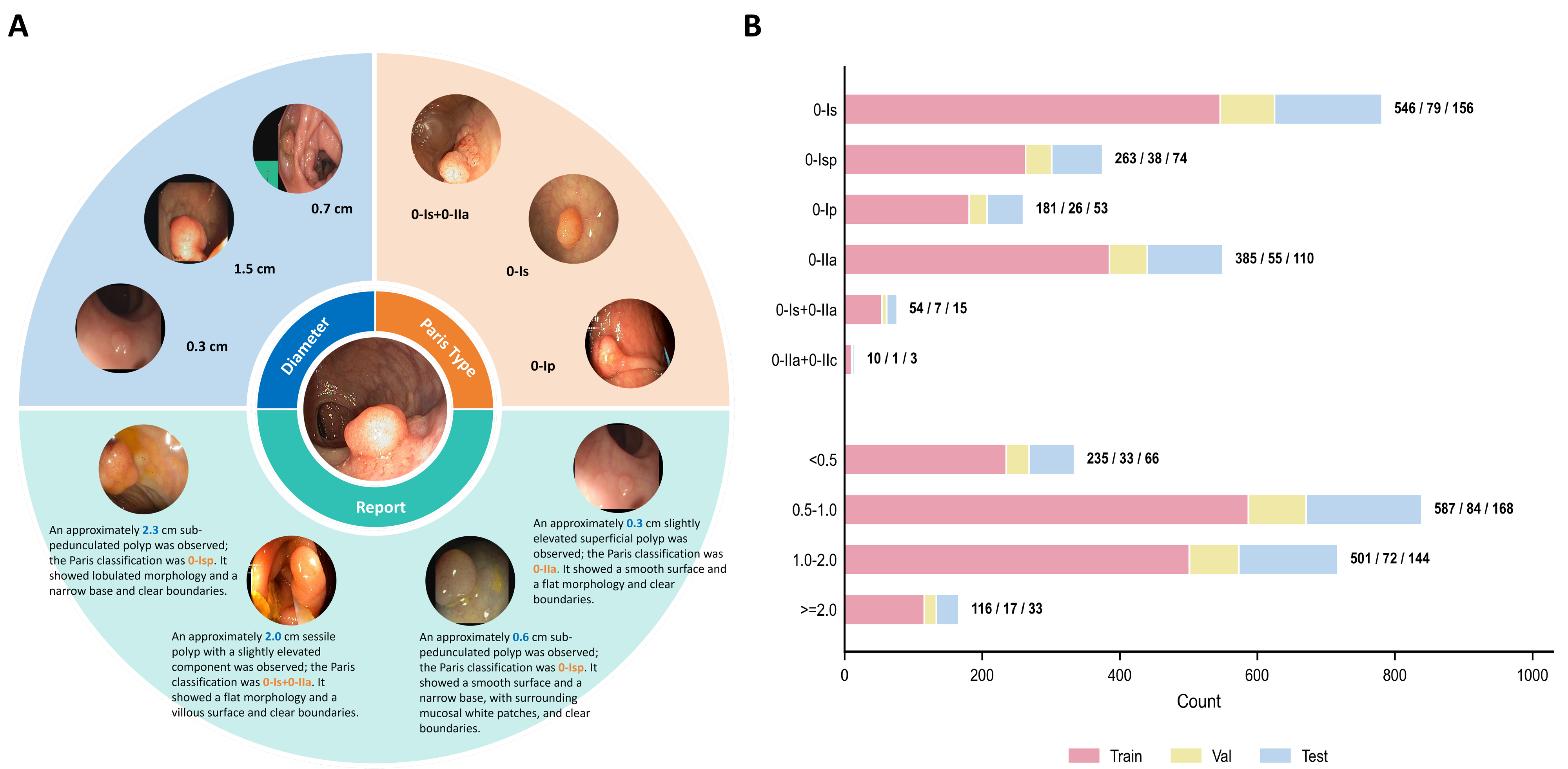}
\caption{Dataset characteristics of the structured polyp-reporting subset. Each image is paired with diameter, Paris type, and a morphology description. These fields are evaluated with numerical, categorical, and language-generation metrics, respectively. The distribution panel summarizes Paris-type and diameter-bin counts across the training, validation, and test splits.}
\label{fig:dataset}
\end{figure*}

\subsection{Training objective}

The ground-truth structured report is appended to $C_q$ as the assistant response and tokenized as $y_{1:T}$. Prompt positions are masked, and the autoregressive loss is applied only to report tokens:
\begin{equation}
\mathcal{L}=-\sum_{t=1}^{T}\log p_{\theta,P}\!\left(y_t\mid y_{<t},C_q\right).
\end{equation}
The VLM parameters $\theta$, EndoFinder encoder, and case bank remain fixed; only $P$ is optimized. Consequently, gradients from the reporting loss update the implicit instruction context without altering the pretrained visual or language representations. At inference, the same sequence structure is retained, the learned $P$ is reused for every query, and the $K$ highest-scoring cases provide explicit transduction context. Prompt templates, optimization settings, and decoding details are provided in the supplementary material.

\section{Experiments}

The experiments compare directly prompted general-purpose VLMs, task-specific predictors, and weight-adaptation methods under the same multi-task reporting setting. We then examine the individual and combined contributions of the two context sources, how retrieval relevance and depth affect performance, whether requesting the three tasks jointly changes their individual performance, and how the two contexts correct errors or handle conflicting evidence in representative cases.

\subsection{Dataset}
\label{sec:task_dataset}

The study subset is derived from the public EndoVL release \cite{fu2026endovl}, which aggregates 2,709 gastrointestinal endoscopic images from nine public datasets. We retained polyp-like lesions with a determinate diameter, a Paris-type label, and a non-empty morphology description. The resulting 2,056 image--report pairs were divided into 1,439 training, 206 validation, and 411 test cases. The split was stratified by the joint label of Paris type and four diameter intervals ($<0.5$, $0.5$--$<1.0$, $1.0$--$<2.0$, and $\geq2.0$~cm), preserving comparable categorical and size distributions across the three subsets (Fig.~\ref{fig:dataset}). Each image--report pair was assigned to exactly one subset. Model optimization and case-bank construction used only the training subset; validation and test cases were used solely as held-out queries, and their reports were never available as retrieval evidence. This separation prevents image- and report-level leakage across the splits. Source-level and label-level split statistics are provided in Supplementary Section 2.

Each report contains diameter, Paris type, and morphological description and is serialized as one JSON response. Figure~\ref{fig:dataset} summarizes the reporting task and label distributions. Diameter is evaluated by mean absolute error (MAE) and threshold agreement at 0.5 and 1.0~cm. These boundaries correspond to commonly used clinical size categories \cite{tsai2011polypSize,usmstf2020}. ACC$>$0.5 records whether the prediction and reference fall on the same side of 0.5~cm; ACC$>$1.0 is defined analogously. Paris type is evaluated by exact-match accuracy and macro precision, recall, and F1, with multi-label combinations treated as complete category values.

For whole-report evaluation, recoverable formatting variants in the predicted JSON objects are first normalized. Each prediction and its reference are then converted into clinical sentences using the same fixed verbalization rule. The resulting sentences contain all three report components: lesion diameter, Paris type, and morphology description. We compute BLEU-4 \cite{bleu2002}, ROUGE-L \cite{rouge2004}, and CIDEr \cite{cider2015} on these complete sentences, complementing the task-specific metrics with sentence-level agreement for the entire report. CIDEr is divided by 10 for display on a 0--1-oriented scale. The normalized JSON objects and verbalized sentences are retained as per-case records. For the general-VLM and main method comparisons, 95\% confidence intervals are estimated from 1,000 case-level bootstrap resamples within each split.

\begin{table*}[t]
\caption{Comparison of general-purpose VLMs and the proposed method on the validation and test sets, with bootstrap 95\% confidence intervals. Bold and underlined values denote the best and second-best distinct point estimates within each split and metric, respectively.}
\label{tab:general_vlm_api}
\centering
\scriptsize
\setlength{\tabcolsep}{2.4pt}
\renewcommand{\arraystretch}{1.35}
\resizebox{\textwidth}{!}{%
\begin{tabular}{@{}llccc@{\hspace{5pt}}cccc@{\hspace{5pt}}ccc@{}}
\toprule
Section & Method & \multicolumn{3}{c}{Diameter} & \multicolumn{4}{c}{Paris type} & \multicolumn{3}{c}{Report generation} \\
 &  & MAE$\downarrow$ & ACC$>$0.5$\uparrow$ & ACC$>$1.0$\uparrow$ & ACC$\uparrow$ & Precision$\uparrow$ & Recall$\uparrow$ & F1$\uparrow$ & BLEU-4$\uparrow$ & ROUGE-L$\uparrow$ & CIDEr$\uparrow$ \\
\midrule
\multirow{6}{*}{Val} & GPT-5.5 & \shortstack{0.354\\{\scriptsize (0.304--0.404)}} & \shortstack{0.684\\{\scriptsize (0.621--0.748)}} & \shortstack{0.665\\{\scriptsize (0.602--0.728)}} & \shortstack{0.296\\{\scriptsize (0.238--0.359)}} & \shortstack{\underline{\smash[b]{0.448}}\\{\scriptsize (0.219--0.505)}} & \shortstack{\underline{\smash[b]{0.265}}\\{\scriptsize (0.203--0.333)}} & \shortstack{\underline{\smash[b]{0.237}}\\{\scriptsize (0.163--0.313)}} & \shortstack{0.386\\{\scriptsize (0.368--0.404)}} & \shortstack{0.651\\{\scriptsize (0.638--0.665)}} & \shortstack{0.063\\{\scriptsize (0.050--0.077)}} \\
 & Claude Sonnet 5 & \shortstack{0.356\\{\scriptsize (0.304--0.408)}} & \shortstack{0.767\\{\scriptsize (0.704--0.821)}} & \shortstack{\underline{\smash[b]{0.728}}\\{\scriptsize (0.670--0.786)}} & \shortstack{0.388\\{\scriptsize (0.320--0.451)}} & \shortstack{0.358\\{\scriptsize (0.181--0.392)}} & \shortstack{0.177\\{\scriptsize (0.142--0.226)}} & \shortstack{0.172\\{\scriptsize (0.113--0.231)}} & \shortstack{0.420\\{\scriptsize (0.401--0.439)}} & \shortstack{0.686\\{\scriptsize (0.673--0.700)}} & \shortstack{0.080\\{\scriptsize (0.064--0.097)}} \\
 & Gemini 3.5 Flash & \shortstack{\underline{\smash[b]{0.314}}\\{\scriptsize (0.269--0.364)}} & \shortstack{\underline{\smash[b]{0.816}}\\{\scriptsize (0.757--0.864)}} & \shortstack{\underline{\smash[b]{0.728}}\\{\scriptsize (0.670--0.787)}} & \shortstack{\underline{\smash[b]{0.403}}\\{\scriptsize (0.335--0.466)}} & \shortstack{0.280\\{\scriptsize (0.238--0.320)}} & \shortstack{0.235\\{\scriptsize (0.192--0.275)}} & \shortstack{0.227\\{\scriptsize (0.183--0.267)}} & \shortstack{\underline{\smash[b]{0.443}}\\{\scriptsize (0.422--0.464)}} & \shortstack{\underline{\smash[b]{0.689}}\\{\scriptsize (0.674--0.704)}} & \shortstack{\underline{\smash[b]{0.086}}\\{\scriptsize (0.073--0.102)}} \\
 & Qwen3.6 Plus & \shortstack{0.374\\{\scriptsize (0.335--0.417)}} & \shortstack{0.748\\{\scriptsize (0.689--0.806)}} & \shortstack{0.670\\{\scriptsize (0.602--0.733)}} & \shortstack{0.325\\{\scriptsize (0.262--0.388)}} & \shortstack{0.198\\{\scriptsize (0.179--0.221)}} & \shortstack{0.138\\{\scriptsize (0.112--0.169)}} & \shortstack{0.121\\{\scriptsize (0.095--0.149)}} & \shortstack{0.389\\{\scriptsize (0.374--0.406)}} & \shortstack{0.660\\{\scriptsize (0.649--0.671)}} & \shortstack{0.060\\{\scriptsize (0.051--0.069)}} \\
 & Frozen VLM & \shortstack{0.569\\{\scriptsize (0.509--0.628)}} & \shortstack{0.534\\{\scriptsize (0.471--0.602)}} & \shortstack{0.408\\{\scriptsize (0.345--0.476)}} & \shortstack{0.214\\{\scriptsize (0.160--0.267)}} & \shortstack{0.240\\{\scriptsize (0.175--0.301)}} & \shortstack{0.200\\{\scriptsize (0.135--0.258)}} & \shortstack{0.149\\{\scriptsize (0.106--0.187)}} & \shortstack{0.350\\{\scriptsize (0.328--0.374)}} & \shortstack{0.619\\{\scriptsize (0.601--0.638)}} & \shortstack{0.065\\{\scriptsize (0.051--0.080)}} \\
 & Ours & \shortstack{\textbf{0.155}\\{\scriptsize (0.120--0.192)}} & \shortstack{\textbf{0.898}\\{\scriptsize (0.854--0.937)}} & \shortstack{\textbf{0.917}\\{\scriptsize (0.879--0.951)}} & \shortstack{\textbf{0.772}\\{\scriptsize (0.714--0.820)}} & \shortstack{\textbf{0.844}\\{\scriptsize (0.642--0.880)}} & \shortstack{\textbf{0.756}\\{\scriptsize (0.530--0.826)}} & \shortstack{\textbf{0.788}\\{\scriptsize (0.560--0.843)}} & \shortstack{\textbf{0.765}\\{\scriptsize (0.732--0.795)}} & \shortstack{\textbf{0.879}\\{\scriptsize (0.860--0.896)}} & \shortstack{\textbf{0.528}\\{\scriptsize (0.476--0.577)}} \\
\midrule
\multirow{6}{*}{Test} & GPT-5.5 & \shortstack{0.373\\{\scriptsize (0.331--0.417)}} & \shortstack{0.708\\{\scriptsize (0.664--0.752)}} & \shortstack{0.679\\{\scriptsize (0.633--0.723)}} & \shortstack{0.304\\{\scriptsize (0.260--0.348)}} & \shortstack{0.214\\{\scriptsize (0.149--0.299)}} & \shortstack{0.159\\{\scriptsize (0.136--0.186)}} & \shortstack{0.140\\{\scriptsize (0.113--0.171)}} & \shortstack{0.385\\{\scriptsize (0.374--0.397)}} & \shortstack{0.652\\{\scriptsize (0.643--0.661)}} & \shortstack{0.057\\{\scriptsize (0.051--0.064)}} \\
 & Claude Sonnet 5 & \shortstack{0.391\\{\scriptsize (0.344--0.441)}} & \shortstack{0.723\\{\scriptsize (0.684--0.769)}} & \shortstack{0.657\\{\scriptsize (0.608--0.703)}} & \shortstack{0.358\\{\scriptsize (0.311--0.401)}} & \shortstack{0.224\\{\scriptsize (0.147--0.302)}} & \shortstack{0.131\\{\scriptsize (0.116--0.147)}} & \shortstack{0.118\\{\scriptsize (0.096--0.139)}} & \shortstack{0.417\\{\scriptsize (0.401--0.432)}} & \shortstack{0.677\\{\scriptsize (0.666--0.688)}} & \shortstack{0.080\\{\scriptsize (0.070--0.092)}} \\
 & Gemini 3.5 Flash & \shortstack{\underline{\smash[b]{0.309}}\\{\scriptsize (0.276--0.345)}} & \shortstack{\underline{\smash[b]{0.781}}\\{\scriptsize (0.740--0.820)}} & \shortstack{\underline{\smash[b]{0.742}}\\{\scriptsize (0.698--0.783)}} & \shortstack{\underline{\smash[b]{0.445}}\\{\scriptsize (0.399--0.494)}} & \shortstack{\underline{\smash[b]{0.380}}\\{\scriptsize (0.280--0.476)}} & \shortstack{\underline{\smash[b]{0.264}}\\{\scriptsize (0.232--0.300)}} & \shortstack{\underline{\smash[b]{0.262}}\\{\scriptsize (0.222--0.304)}} & \shortstack{\underline{\smash[b]{0.453}}\\{\scriptsize (0.438--0.467)}} & \shortstack{\underline{\smash[b]{0.689}}\\{\scriptsize (0.680--0.699)}} & \shortstack{\underline{\smash[b]{0.092}}\\{\scriptsize (0.082--0.103)}} \\
 & Qwen3.6 Plus & \shortstack{0.375\\{\scriptsize (0.344--0.407)}} & \shortstack{\underline{\smash[b]{0.781}}\\{\scriptsize (0.742--0.820)}} & \shortstack{0.698\\{\scriptsize (0.657--0.742)}} & \shortstack{0.324\\{\scriptsize (0.282--0.367)}} & \shortstack{0.189\\{\scriptsize (0.158--0.223)}} & \shortstack{0.165\\{\scriptsize (0.144--0.188)}} & \shortstack{0.143\\{\scriptsize (0.122--0.164)}} & \shortstack{0.386\\{\scriptsize (0.375--0.398)}} & \shortstack{0.655\\{\scriptsize (0.647--0.664)}} & \shortstack{0.059\\{\scriptsize (0.053--0.066)}} \\
 & Frozen VLM & \shortstack{0.534\\{\scriptsize (0.488--0.582)}} & \shortstack{0.564\\{\scriptsize (0.516--0.611)}} & \shortstack{0.479\\{\scriptsize (0.431--0.528)}} & \shortstack{0.197\\{\scriptsize (0.161--0.234)}} & \shortstack{0.215\\{\scriptsize (0.167--0.256)}} & \shortstack{0.200\\{\scriptsize (0.163--0.234)}} & \shortstack{0.135\\{\scriptsize (0.107--0.162)}} & \shortstack{0.357\\{\scriptsize (0.341--0.373)}} & \shortstack{0.625\\{\scriptsize (0.612--0.638)}} & \shortstack{0.069\\{\scriptsize (0.058--0.081)}} \\
 & Ours & \shortstack{\textbf{0.200}\\{\scriptsize (0.168--0.234)}} & \shortstack{\textbf{0.898}\\{\scriptsize (0.866--0.927)}} & \shortstack{\textbf{0.878}\\{\scriptsize (0.847--0.905)}} & \shortstack{\textbf{0.698}\\{\scriptsize (0.650--0.740)}} & \shortstack{\textbf{0.791}\\{\scriptsize (0.626--0.835)}} & \shortstack{\textbf{0.716}\\{\scriptsize (0.555--0.769)}} & \shortstack{\textbf{0.744}\\{\scriptsize (0.577--0.786)}} & \shortstack{\textbf{0.730}\\{\scriptsize (0.704--0.753)}} & \shortstack{\textbf{0.858}\\{\scriptsize (0.843--0.871)}} & \shortstack{\textbf{0.491}\\{\scriptsize (0.452--0.523)}} \\
\bottomrule
\end{tabular}%
}
\end{table*}

\begin{table*}[t]
\caption{Comparison with retrieval-based and single-task predictors on validation and test sets. Lower MAE is better for diameter; higher ACC is better for Paris type. Bold and underlined values denote the best and second-best distinct point estimates, respectively. A dash indicates that a single-task predictor does not produce the other task's output.}
\label{tab:domain_specialized_baselines}
\centering
\begin{tabular}{@{}llcccc@{}}
\toprule
Method & Output scope & Val MAE$\downarrow$ & Val ACC$\uparrow$ & Test MAE$\downarrow$ & Test ACC$\uparrow$ \\
\midrule
EndoFinder retrieval baseline & Task-specific & 0.321 & 0.655 & 0.311 & 0.638 \\
Supervised ViT Paris classifier & Single-task & -- & \underline{\smash[b]{0.670}} & -- & \underline{\smash[b]{0.657}} \\
Supervised ViT diameter regressor & Single-task & \underline{\smash[b]{0.279}} & -- & \underline{\smash[b]{0.306}} & -- \\
\midrule
Ours & Unified report & \textbf{0.155} & \textbf{0.772} & \textbf{0.200} & \textbf{0.698} \\
\bottomrule
\end{tabular}
\end{table*}

\subsection{Implementation details}

All local experiments used Qwen3.5-9B \cite{qwen35_2026} with bfloat16 computation on one NVIDIA A100 80GB GPU. Images were supplied at their native resolution and processed by the model's variable-resolution visual tokenizer \cite{qwen2vl2024}, which uses $16\times16$ patches and a spatial merge factor of 2. The retrieval bank contains 1,024-dimensional L2-normalized EndoFinder CLS representations from the training images. Validation and test images were used only as retrieval queries.

The Prompt tuning baseline inserts 128 trainable continuous tokens into the input prompt without retrieved cases, corresponding to 0.524~M trainable parameters. The continuous tokens were optimized for 20 epochs with AdamW, a peak learning rate of $10^{-4}$, weight decay of 0.1, an effective batch size of 16, OneCycle scheduling, and gradient clipping at 1.0. Prompt positions were masked from the autoregressive loss. Unless otherwise stated, inference uses the five highest-ranked training cases and greedy decoding with at most 128 new tokens.

The directly prompted models were GPT-5.5\cite{openaiGpt55Docs} , Claude Sonnet 5 \cite{anthropicSonnet5Docs}, Gemini 3.5 Flash \cite{googleGemini35FlashDocs}, and Qwen3.6 Plus \cite{qwen36PlusDocs}. Each model received one query image, identical field definitions, Paris categories, and output schema, without retrieved cases or specialist examples.

All adaptation methods use the same Qwen3.5-9B backbone and evaluation pipeline. Finetune updates the VLM's 40.119~M-parameter visual--language merger while keeping the remaining backbone frozen. LoRA inserts rank-8 attention adapters and optimizes 9.056~M parameters. Direct inference and MM-RAG require no optimization, whereas Prompt tuning and Ours optimize only the continuous specialist tokens. The exact trainable locations and optimization settings for the weight-adaptation baselines are provided in Supplementary Section 3 and Supplementary Fig. S1. Task-specific comparisons use the EndoFinder retrieval estimate, a supervised ViT Paris classifier, and a supervised ViT diameter regressor.

For the context-quality study, random cases are sampled without reference to the query, EndoFinder cases follow cosine similarity in the fixed CLS space, and a label oracle provides an analytical upper bound. The oracle score combines Paris-type agreement, diameter proximity, and morphology-term overlap with weights 0.40, 0.35, and 0.25, respectively. It is used only to measure the headroom available from improved evidence selection.

\begin{table*}[!t]
\caption{Main comparison on validation and test sets with bootstrap 95\% confidence intervals. Bold and underlined values denote the best and second-best distinct point estimates within each split and metric, respectively.}
\label{tab:main_results}
\centering
\scriptsize
\setlength{\tabcolsep}{2.4pt}
\renewcommand{\arraystretch}{1.85}
\resizebox{\textwidth}{!}{%
\begin{tabular}{@{}llccc@{\hspace{5pt}}cccc@{\hspace{5pt}}ccc@{}}
\toprule
Section & Method & \multicolumn{3}{c}{Diameter} & \multicolumn{4}{c}{Paris type} & \multicolumn{3}{c}{Report generation} \\
 &  & MAE$\downarrow$ & ACC$>$0.5$\uparrow$ & ACC$>$1.0$\uparrow$ & ACC$\uparrow$ & Precision$\uparrow$ & Recall$\uparrow$ & F1$\uparrow$ & BLEU-4$\uparrow$ & ROUGE-L$\uparrow$ & CIDEr$\uparrow$ \\
\midrule
\multirow{6}{*}{Val} & Frozen VLM & \shortstack{0.569\\{\scriptsize (0.509--0.628)}} & \shortstack{0.534\\{\scriptsize (0.471--0.602)}} & \shortstack{0.408\\{\scriptsize (0.345--0.476)}} & \shortstack{0.214\\{\scriptsize (0.160--0.267)}} & \shortstack{0.240\\{\scriptsize (0.175--0.301)}} & \shortstack{0.200\\{\scriptsize (0.135--0.258)}} & \shortstack{0.149\\{\scriptsize (0.106--0.187)}} & \shortstack{0.350\\{\scriptsize (0.328--0.374)}} & \shortstack{0.619\\{\scriptsize (0.601--0.638)}} & \shortstack{0.065\\{\scriptsize (0.051--0.080)}} \\
 & Finetune & \shortstack{0.208\\{\scriptsize (0.169--0.250)}} & \shortstack{\textbf{0.908}\\{\scriptsize (0.869--0.942)}} & \shortstack{\underline{\smash[b]{0.908}}\\{\scriptsize (0.864--0.947)}} & \shortstack{\underline{\smash[b]{0.718}}\\{\scriptsize (0.655--0.777)}} & \shortstack{0.751\\{\scriptsize (0.534--0.818)}} & \shortstack{\underline{\smash[b]{0.753}}\\{\scriptsize (0.529--0.817)}} & \shortstack{\underline{\smash[b]{0.740}}\\{\scriptsize (0.514--0.797)}} & \shortstack{\underline{\smash[b]{0.720}}\\{\scriptsize (0.687--0.752)}} & \shortstack{\underline{\smash[b]{0.856}}\\{\scriptsize (0.839--0.875)}} & \shortstack{\underline{\smash[b]{0.453}}\\{\scriptsize (0.409--0.501)}} \\
 & LoRA & \shortstack{\underline{\smash[b]{0.195}}\\{\scriptsize (0.162--0.234)}} & \shortstack{0.879\\{\scriptsize (0.830--0.922)}} & \shortstack{0.893\\{\scriptsize (0.850--0.932)}} & \shortstack{0.699\\{\scriptsize (0.631--0.762)}} & \shortstack{\underline{\smash[b]{0.758}}\\{\scriptsize (0.535--0.829)}} & \shortstack{0.725\\{\scriptsize (0.499--0.792)}} & \shortstack{0.727\\{\scriptsize (0.503--0.787)}} & \shortstack{0.689\\{\scriptsize (0.658--0.721)}} & \shortstack{0.837\\{\scriptsize (0.819--0.855)}} & \shortstack{0.404\\{\scriptsize (0.361--0.448)}} \\
 & MM-RAG & \shortstack{0.216\\{\scriptsize (0.172--0.266)}} & \shortstack{0.859\\{\scriptsize (0.811--0.903)}} & \shortstack{0.888\\{\scriptsize (0.845--0.927)}} & \shortstack{0.641\\{\scriptsize (0.573--0.704)}} & \shortstack{0.689\\{\scriptsize (0.570--0.747)}} & \shortstack{0.678\\{\scriptsize (0.537--0.752)}} & \shortstack{0.662\\{\scriptsize (0.525--0.718)}} & \shortstack{0.694\\{\scriptsize (0.659--0.729)}} & \shortstack{0.840\\{\scriptsize (0.821--0.860)}} & \shortstack{0.434\\{\scriptsize (0.383--0.484)}} \\
 & Prompt tuning & \shortstack{0.244\\{\scriptsize (0.203--0.294)}} & \shortstack{0.864\\{\scriptsize (0.820--0.908)}} & \shortstack{0.869\\{\scriptsize (0.820--0.913)}} & \shortstack{0.641\\{\scriptsize (0.568--0.704)}} & \shortstack{0.505\\{\scriptsize (0.396--0.609)}} & \shortstack{0.463\\{\scriptsize (0.391--0.541)}} & \shortstack{0.476\\{\scriptsize (0.391--0.551)}} & \shortstack{0.668\\{\scriptsize (0.633--0.700)}} & \shortstack{0.826\\{\scriptsize (0.807--0.844)}} & \shortstack{0.387\\{\scriptsize (0.342--0.432)}} \\
 & Ours & \shortstack{\textbf{0.155}\\{\scriptsize (0.120--0.192)}} & \shortstack{\underline{\smash[b]{0.898}}\\{\scriptsize (0.854--0.937)}} & \shortstack{\textbf{0.917}\\{\scriptsize (0.879--0.951)}} & \shortstack{\textbf{0.772}\\{\scriptsize (0.714--0.820)}} & \shortstack{\textbf{0.844}\\{\scriptsize (0.642--0.880)}} & \shortstack{\textbf{0.756}\\{\scriptsize (0.530--0.826)}} & \shortstack{\textbf{0.788}\\{\scriptsize (0.560--0.843)}} & \shortstack{\textbf{0.765}\\{\scriptsize (0.732--0.795)}} & \shortstack{\textbf{0.879}\\{\scriptsize (0.860--0.896)}} & \shortstack{\textbf{0.528}\\{\scriptsize (0.476--0.577)}} \\
\midrule
\multirow{6}{*}{Test} & Frozen VLM & \shortstack{0.534\\{\scriptsize (0.488--0.582)}} & \shortstack{0.564\\{\scriptsize (0.516--0.611)}} & \shortstack{0.479\\{\scriptsize (0.431--0.528)}} & \shortstack{0.197\\{\scriptsize (0.161--0.234)}} & \shortstack{0.215\\{\scriptsize (0.167--0.256)}} & \shortstack{0.200\\{\scriptsize (0.163--0.234)}} & \shortstack{0.135\\{\scriptsize (0.107--0.162)}} & \shortstack{0.357\\{\scriptsize (0.341--0.373)}} & \shortstack{0.625\\{\scriptsize (0.612--0.638)}} & \shortstack{0.069\\{\scriptsize (0.058--0.081)}} \\
 & Finetune & \shortstack{\underline{\smash[b]{0.203}}\\{\scriptsize (0.174--0.236)}} & \shortstack{\textbf{0.903}\\{\scriptsize (0.873--0.930)}} & \shortstack{\textbf{0.883}\\{\scriptsize (0.852--0.912)}} & \shortstack{\underline{\smash[b]{0.664}}\\{\scriptsize (0.618--0.708)}} & \shortstack{\underline{\smash[b]{0.717}}\\{\scriptsize (0.531--0.767)}} & \shortstack{0.629\\{\scriptsize (0.495--0.723)}} & \shortstack{0.659\\{\scriptsize (0.503--0.725)}} & \shortstack{\underline{\smash[b]{0.697}}\\{\scriptsize (0.673--0.720)}} & \shortstack{\underline{\smash[b]{0.843}}\\{\scriptsize (0.830--0.856)}} & \shortstack{\underline{\smash[b]{0.434}}\\{\scriptsize (0.400--0.468)}} \\
 & LoRA & \shortstack{0.226\\{\scriptsize (0.195--0.261)}} & \shortstack{0.883\\{\scriptsize (0.852--0.912)}} & \shortstack{\textbf{0.883}\\{\scriptsize (0.852--0.912)}} & \shortstack{0.650\\{\scriptsize (0.603--0.691)}} & \shortstack{0.676\\{\scriptsize (0.552--0.760)}} & \shortstack{0.678\\{\scriptsize (0.513--0.731)}} & \shortstack{\underline{\smash[b]{0.663}}\\{\scriptsize (0.517--0.724)}} & \shortstack{0.673\\{\scriptsize (0.650--0.697)}} & \shortstack{0.826\\{\scriptsize (0.813--0.840)}} & \shortstack{0.396\\{\scriptsize (0.365--0.431)}} \\
 & MM-RAG & \shortstack{0.235\\{\scriptsize (0.198--0.277)}} & \shortstack{0.878\\{\scriptsize (0.847--0.908)}} & \shortstack{0.869\\{\scriptsize (0.835--0.900)}} & \shortstack{0.635\\{\scriptsize (0.589--0.679)}} & \shortstack{0.700\\{\scriptsize (0.638--0.745)}} & \shortstack{\underline{\smash[b]{0.682}}\\{\scriptsize (0.611--0.734)}} & \shortstack{\underline{\smash[b]{0.663}}\\{\scriptsize (0.591--0.707)}} & \shortstack{0.683\\{\scriptsize (0.659--0.707)}} & \shortstack{0.830\\{\scriptsize (0.816--0.844)}} & \shortstack{0.423\\{\scriptsize (0.388--0.459)}} \\
 & Prompt tuning & \shortstack{0.281\\{\scriptsize (0.241--0.323)}} & \shortstack{0.869\\{\scriptsize (0.837--0.900)}} & \shortstack{0.844\\{\scriptsize (0.808--0.876)}} & \shortstack{0.625\\{\scriptsize (0.579--0.669)}} & \shortstack{0.524\\{\scriptsize (0.467--0.575)}} & \shortstack{0.511\\{\scriptsize (0.454--0.563)}} & \shortstack{0.513\\{\scriptsize (0.458--0.557)}} & \shortstack{0.661\\{\scriptsize (0.637--0.683)}} & \shortstack{0.822\\{\scriptsize (0.809--0.835)}} & \shortstack{0.379\\{\scriptsize (0.347--0.409)}} \\
 & Ours & \shortstack{\textbf{0.200}\\{\scriptsize (0.168--0.234)}} & \shortstack{\underline{\smash[b]{0.898}}\\{\scriptsize (0.866--0.927)}} & \shortstack{\underline{\smash[b]{0.878}}\\{\scriptsize (0.847--0.905)}} & \shortstack{\textbf{0.698}\\{\scriptsize (0.650--0.740)}} & \shortstack{\textbf{0.791}\\{\scriptsize (0.626--0.835)}} & \shortstack{\textbf{0.716}\\{\scriptsize (0.555--0.769)}} & \shortstack{\textbf{0.744}\\{\scriptsize (0.577--0.786)}} & \shortstack{\textbf{0.730}\\{\scriptsize (0.704--0.753)}} & \shortstack{\textbf{0.858}\\{\scriptsize (0.843--0.871)}} & \shortstack{\textbf{0.491}\\{\scriptsize (0.452--0.523)}} \\
\bottomrule
\end{tabular}%
}
\end{table*}

\begin{figure*}[t]
\centering
\includegraphics[width=\textwidth]{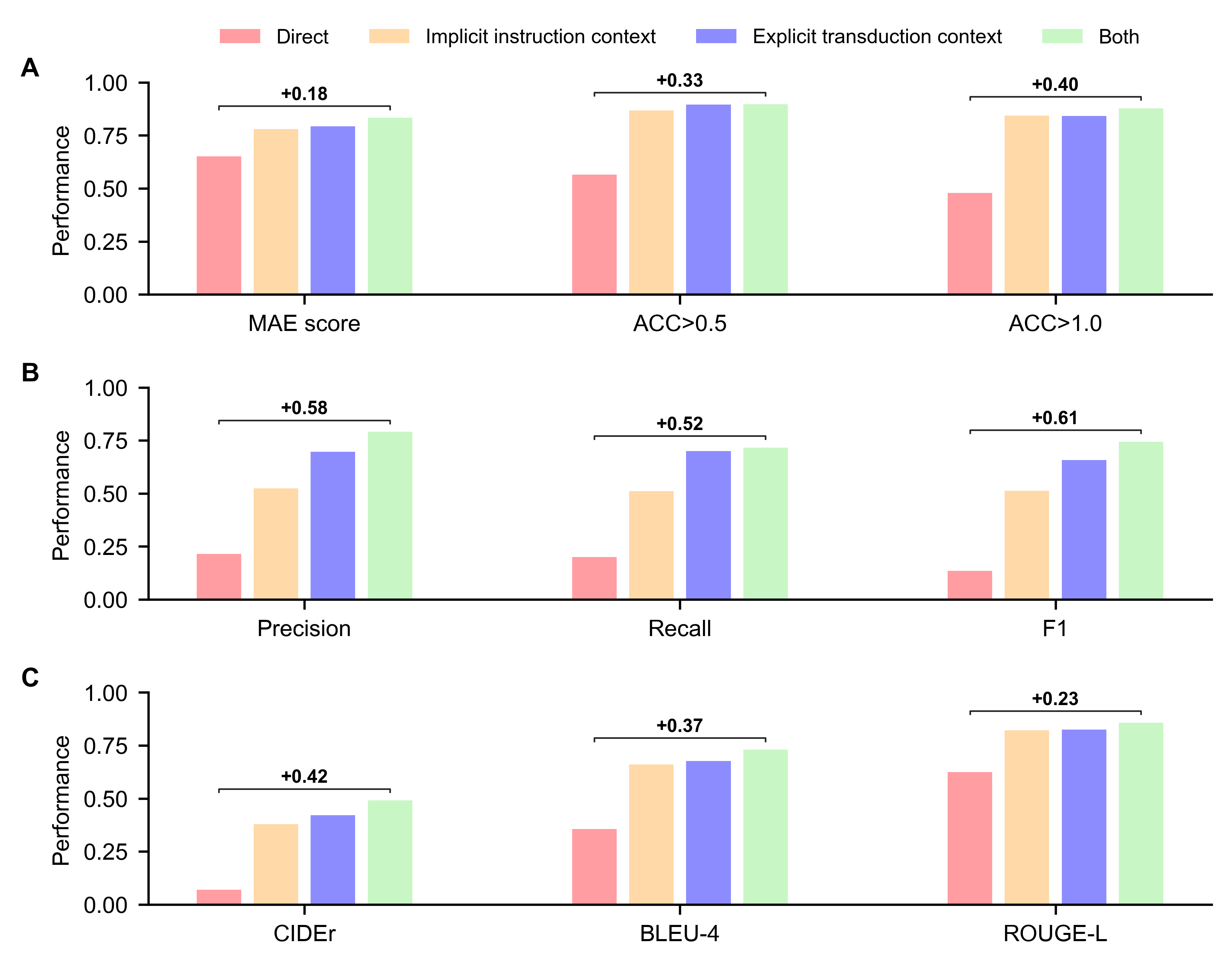}
\caption{Effect of explicit transduction context and implicit instruction context. Direct denotes frozen-VLM inference without either context; the remaining settings use implicit instruction context, explicit transduction context, or both. The three rows evaluate diameter estimation, Paris-type classification, and report generation after converting structured JSON outputs into the fixed clinical sentence template. For display in panel A, MAE is transformed to \(1/(1+\mathrm{MAE})\), so all bars use a higher-is-better orientation. The combined setting yielded the highest overall values across the three metric groups.}
\label{fig:memory_context_effects}
\end{figure*}

\begin{table*}[t]
\caption{Comparison of context configurations with five multimodal context cases. Random MM-RAG denotes MM-RAG supplied with randomly sampled training image--report pairs instead of EndoFinder-retrieved cases; the number and modalities of the context cases remain unchanged. Bold and underlined values denote the best and second-best distinct point estimates within each split and metric, respectively.}
\label{tab:context_length_control}
\centering
\scriptsize
\setlength{\tabcolsep}{1.8pt}
\renewcommand{\arraystretch}{1.15}
\resizebox{\textwidth}{!}{%
\begin{tabular}{@{}llccc@{\hspace{5pt}}cccc@{\hspace{5pt}}ccc@{}}
\toprule
Section & Method & \multicolumn{3}{c}{Diameter} & \multicolumn{4}{c}{Paris type} & \multicolumn{3}{c}{Report generation} \\
 &  & MAE$\downarrow$ & ACC$>$0.5$\uparrow$ & ACC$>$1.0$\uparrow$ & ACC$\uparrow$ & Precision$\uparrow$ & Recall$\uparrow$ & F1$\uparrow$ & BLEU-4$\uparrow$ & ROUGE-L$\uparrow$ & CIDEr$\uparrow$ \\
\midrule
Val & MM-RAG with random cases & 0.365 & 0.806 & 0.786 & 0.369 & 0.415 & 0.272 & 0.216 & 0.502 & 0.730 & 0.143 \\
 & Prompt tuning + random MM-RAG & 0.320 & 0.840 & 0.811 & 0.597 & 0.576 & 0.384 & 0.411 & 0.608 & 0.786 & 0.272 \\
 & MM-RAG & \underline{\smash[b]{0.216}} & 0.859 & \underline{\smash[b]{0.888}} & \underline{\smash[b]{0.641}} & \underline{\smash[b]{0.689}} & \underline{\smash[b]{0.678}} & \underline{\smash[b]{0.662}} & \underline{\smash[b]{0.694}} & \underline{\smash[b]{0.840}} & \underline{\smash[b]{0.434}} \\
 & Prompt tuning & 0.244 & \underline{\smash[b]{0.864}} & 0.869 & \underline{\smash[b]{0.641}} & 0.505 & 0.463 & 0.476 & 0.668 & 0.826 & 0.387 \\
 & Ours & \textbf{0.155} & \textbf{0.898} & \textbf{0.917} & \textbf{0.772} & \textbf{0.844} & \textbf{0.756} & \textbf{0.788} & \textbf{0.765} & \textbf{0.879} & \textbf{0.528} \\
\midrule
Test & MM-RAG with random cases & 0.400 & 0.825 & 0.771 & 0.377 & 0.351 & 0.255 & 0.204 & 0.512 & 0.735 & 0.176 \\
 & Prompt tuning + random MM-RAG & 0.337 & 0.839 & 0.810 & 0.523 & 0.356 & 0.316 & 0.321 & 0.585 & 0.775 & 0.262 \\
 & MM-RAG & \underline{\smash[b]{0.235}} & \underline{\smash[b]{0.878}} & \underline{\smash[b]{0.869}} & \underline{\smash[b]{0.635}} & \underline{\smash[b]{0.700}} & \underline{\smash[b]{0.682}} & \underline{\smash[b]{0.663}} & \underline{\smash[b]{0.683}} & \underline{\smash[b]{0.830}} & \underline{\smash[b]{0.423}} \\
 & Prompt tuning & 0.281 & 0.869 & 0.844 & 0.625 & 0.524 & 0.511 & 0.513 & 0.661 & 0.822 & 0.379 \\
 & Ours & \textbf{0.200} & \textbf{0.898} & \textbf{0.878} & \textbf{0.698} & \textbf{0.791} & \textbf{0.716} & \textbf{0.744} & \textbf{0.730} & \textbf{0.858} & \textbf{0.491} \\
\bottomrule
\end{tabular}%
}
\end{table*}

\subsection{Benchmark of general-purpose VLMs}

General-purpose VLMs followed the reporting instruction, but their diameter and Paris-type predictions remained unreliable (Table~\ref{tab:general_vlm_api}). Gemini 3.5 Flash was the strongest directly prompted API model on the test set, with a diameter MAE of 0.309 and Paris-type accuracy of 0.445, but its Paris macro-F1 was only 0.262. Direct frozen-VLM inference was weaker, with a diameter MAE of 0.534 and Paris-type accuracy of 0.197. The validation results showed the same overall ordering.

The metric profile separates fluent reporting from accurate specialist interpretation. The strongest direct API result reached 0.689 ROUGE-L on the test set, although fewer than half of its Paris labels were exactly correct. These models often reproduced the expected sentence structure and common morphology terms while making errors in quantitative or categorical fields. A single language-generation score would therefore obscure clinically important failures.

The proposed context-fusion framework reduced test diameter MAE to 0.200, increased Paris accuracy and macro-F1 to 0.698 and 0.744, and raised ROUGE-L to 0.858 (Table~\ref{tab:general_vlm_api}). Together, these results show that context fusion improves both clinically verifiable fields and overall report quality.

\subsection{Comparison with specialist adaptation baselines}

The unified framework also exceeded the task-specific predictors (Table~\ref{tab:domain_specialized_baselines}). On the test set, its diameter MAE was 0.200, compared with 0.306 for the supervised ViT regressor, while its Paris accuracy was 0.698, compared with 0.657 for the supervised ViT classifier. Unlike these single-task models, the framework generates diameter, Paris type, and morphology in one response.

Weight adaptation was competitive but did not dominate across the complete report (Table~\ref{tab:main_results}). Finetune achieved the highest test ACC$>$0.5 and tied with LoRA for the highest ACC$>$1.0, indicating strong separation around the two diameter thresholds. Ours achieved a slightly lower continuous MAE than Finetune (0.200 versus 0.203), higher Paris accuracy (0.698 versus 0.664), and higher ROUGE-L (0.858 versus 0.843). LoRA reached 0.226 MAE, 0.650 Paris accuracy, and 0.826 ROUGE-L. The context-based framework therefore provided the strongest overall balance across numerical, categorical, and generative outputs.

This balance required substantially fewer trainable parameters. As shown in Fig.~\ref{fig:parameter_efficiency} and Table~\ref{tab:main_results}, Ours optimized 0.524~M parameters, whereas LoRA and Finetune required 9.056~M and 40.119~M, respectively (17.3 and 76.6 times as many).

We further examined Paris-type cases that Finetune classified incorrectly. Among the 61 such cases for which the top-1 retrieved report contained the target Paris type, Ours corrected 43 (70.5\%). In the corresponding stratum of 201 cases that Finetune classified correctly, Ours changed 14 predictions to errors (7.0\%).

\begin{center}
\centering
\includegraphics[width=\columnwidth]{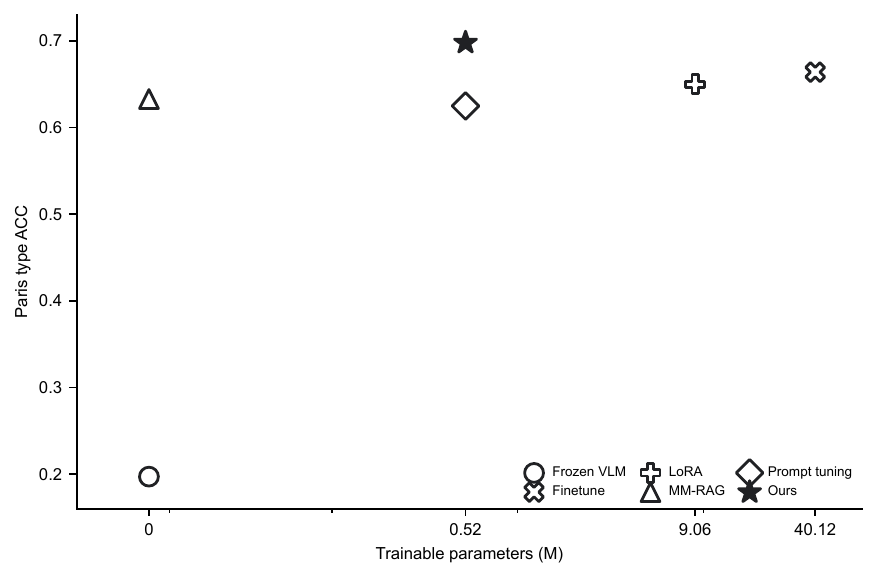}
\captionof{figure}{Parameter efficiency on the test set. Paris-type accuracy is plotted against the number of trainable parameters. Direct inference and MM-RAG leave the backbone unchanged and use no trainable parameters; Prompt tuning and Ours optimize only 0.524~M specialist-token parameters. The proposed combination attains the highest Paris-type accuracy with substantially fewer trainable parameters than LoRA and visual--language merger fine-tuning.}
\label{fig:parameter_efficiency}
\end{center}

\begin{table*}[t]
\caption{Comparison of report-generation performance across context sizes and evidence sources using internal-context multimodal RAG. At each \(K\), random, EndoFinder, and oracle evidence are evaluated with the same number of retrieved cases. Bold and underlined values denote the best and second-best distinct point estimates within each split, \(K\), and metric, respectively.}
\label{tab:context_quality_topk}
\centering
\scriptsize
\setlength{\tabcolsep}{1.55pt}
\renewcommand{\arraystretch}{1.10}
\resizebox{\textwidth}{!}{%
\begin{tabular}{@{}lllccc@{\hspace{4pt}}cccc@{\hspace{4pt}}ccc@{\hspace{4pt}}cc@{}}
\toprule
Section & \(K\) & Evidence source & \multicolumn{3}{c}{Diameter} & \multicolumn{4}{c}{Paris type} & \multicolumn{3}{c}{Report generation} & \multicolumn{2}{c}{Evidence coverage} \\
 &  &  & MAE$\downarrow$ & ACC$>$0.5$\uparrow$ & ACC$>$1.0$\uparrow$ & ACC$\uparrow$ & Precision$\uparrow$ & Recall$\uparrow$ & F1$\uparrow$ & BLEU-4$\uparrow$ & ROUGE-L$\uparrow$ & CIDEr$\uparrow$ & Paris hit@\(K\)$\uparrow$ & Size hit@\(K\)$\uparrow$ \\
\midrule
Val & \multirow{3}{*}{1} & Random cases & 0.311 & 0.816 & \underline{\smash[b]{0.801}} & 0.553 & 0.571 & 0.329 & 0.346 & 0.586 & 0.777 & 0.233 & 0.252 & 0.558 \\
 &  & EndoFinder & \underline{\smash[b]{0.206}} & \underline{\smash[b]{0.850}} & \textbf{0.879} & \underline{\smash[b]{0.723}} & \underline{\smash[b]{0.869}} & \underline{\smash[b]{0.724}} & \underline{\smash[b]{0.769}} & \underline{\smash[b]{0.717}} & \underline{\smash[b]{0.850}} & \underline{\smash[b]{0.441}} & \underline{\smash[b]{0.655}} & \underline{\smash[b]{0.825}} \\
 &  & Oracle & \textbf{0.168} & \textbf{0.908} & \textbf{0.879} & \textbf{0.816} & \textbf{0.934} & \textbf{0.798} & \textbf{0.846} & \textbf{0.782} & \textbf{0.881} & \textbf{0.551} & \textbf{1.000} & \textbf{0.990} \\
\addlinespace[1pt]
 & \multirow{3}{*}{3} & Random cases & 0.318 & 0.820 & 0.825 & 0.597 & 0.580 & 0.372 & 0.398 & 0.601 & 0.784 & 0.258 & 0.578 & 0.859 \\
 &  & EndoFinder & \underline{\smash[b]{0.163}} & \underline{\smash[b]{0.874}} & \underline{\smash[b]{0.908}} & \underline{\smash[b]{0.738}} & \underline{\smash[b]{0.817}} & \underline{\smash[b]{0.728}} & \underline{\smash[b]{0.758}} & \underline{\smash[b]{0.751}} & \underline{\smash[b]{0.870}} & \underline{\smash[b]{0.513}} & \underline{\smash[b]{0.845}} & \underline{\smash[b]{0.932}} \\
 &  & Oracle & \textbf{0.117} & \textbf{0.937} & \textbf{0.927} & \textbf{0.903} & \textbf{0.954} & \textbf{0.922} & \textbf{0.935} & \textbf{0.868} & \textbf{0.933} & \textbf{0.701} & \textbf{1.000} & \textbf{1.000} \\
\addlinespace[1pt]
 & \multirow{3}{*}{5} & Random cases & 0.320 & 0.840 & 0.811 & 0.597 & 0.576 & 0.384 & 0.411 & 0.608 & 0.786 & 0.272 & 0.757 & 0.951 \\
 &  & EndoFinder & \underline{\smash[b]{0.155}} & \underline{\smash[b]{0.898}} & \underline{\smash[b]{0.917}} & \underline{\smash[b]{0.772}} & \underline{\smash[b]{0.844}} & \underline{\smash[b]{0.756}} & \underline{\smash[b]{0.788}} & \underline{\smash[b]{0.765}} & \underline{\smash[b]{0.879}} & \underline{\smash[b]{0.528}} & \underline{\smash[b]{0.908}} & \underline{\smash[b]{0.961}} \\
 &  & Oracle & \textbf{0.114} & \textbf{0.932} & \textbf{0.937} & \textbf{0.903} & \textbf{0.956} & \textbf{0.898} & \textbf{0.923} & \textbf{0.872} & \textbf{0.936} & \textbf{0.709} & \textbf{1.000} & \textbf{1.000} \\
\addlinespace[1pt]
 & \multirow{3}{*}{7} & Random cases & 0.306 & 0.811 & 0.816 & 0.617 & 0.600 & 0.400 & 0.429 & 0.618 & 0.796 & 0.286 & 0.835 & \underline{\smash[b]{0.976}} \\
 &  & EndoFinder & \underline{\smash[b]{0.164}} & \underline{\smash[b]{0.893}} & \underline{\smash[b]{0.893}} & \underline{\smash[b]{0.752}} & \underline{\smash[b]{0.844}} & \underline{\smash[b]{0.724}} & \underline{\smash[b]{0.759}} & \underline{\smash[b]{0.749}} & \underline{\smash[b]{0.871}} & \underline{\smash[b]{0.501}} & \underline{\smash[b]{0.956}} & 0.966 \\
 &  & Oracle & \textbf{0.110} & \textbf{0.932} & \textbf{0.927} & \textbf{0.917} & \textbf{0.958} & \textbf{0.911} & \textbf{0.931} & \textbf{0.869} & \textbf{0.935} & \textbf{0.702} & \textbf{1.000} & \textbf{1.000} \\
\midrule
Test & \multirow{3}{*}{1} & Random cases & 0.344 & 0.847 & 0.800 & 0.479 & 0.343 & 0.258 & 0.250 & 0.566 & 0.762 & 0.229 & 0.285 & 0.613 \\
 &  & EndoFinder & \underline{\smash[b]{0.231}} & \underline{\smash[b]{0.903}} & \underline{\smash[b]{0.871}} & \underline{\smash[b]{0.647}} & \underline{\smash[b]{0.812}} & \underline{\smash[b]{0.652}} & \underline{\smash[b]{0.699}} & \underline{\smash[b]{0.693}} & \underline{\smash[b]{0.837}} & \underline{\smash[b]{0.428}} & \underline{\smash[b]{0.637}} & \underline{\smash[b]{0.839}} \\
 &  & Oracle & \textbf{0.182} & \textbf{0.912} & \textbf{0.888} & \textbf{0.723} & \textbf{0.869} & \textbf{0.737} & \textbf{0.780} & \textbf{0.738} & \textbf{0.858} & \textbf{0.502} & \textbf{1.000} & \textbf{0.985} \\
\addlinespace[1pt]
 & \multirow{3}{*}{3} & Random cases & 0.358 & 0.849 & 0.791 & 0.504 & 0.347 & 0.294 & 0.298 & 0.574 & 0.769 & 0.242 & 0.596 & 0.878 \\
 &  & EndoFinder & \underline{\smash[b]{0.204}} & \underline{\smash[b]{0.895}} & \underline{\smash[b]{0.878}} & \underline{\smash[b]{0.703}} & \underline{\smash[b]{0.822}} & \underline{\smash[b]{0.721}} & \underline{\smash[b]{0.758}} & \underline{\smash[b]{0.732}} & \underline{\smash[b]{0.859}} & \underline{\smash[b]{0.489}} & \underline{\smash[b]{0.820}} & \underline{\smash[b]{0.939}} \\
 &  & Oracle & \textbf{0.130} & \textbf{0.942} & \textbf{0.920} & \textbf{0.859} & \textbf{0.928} & \textbf{0.882} & \textbf{0.901} & \textbf{0.839} & \textbf{0.916} & \textbf{0.669} & \textbf{1.000} & \textbf{0.998} \\
\addlinespace[1pt]
 & \multirow{3}{*}{5} & Random cases & 0.337 & 0.839 & 0.810 & 0.523 & 0.356 & 0.316 & 0.321 & 0.585 & 0.775 & 0.262 & 0.762 & 0.944 \\
 &  & EndoFinder & \underline{\smash[b]{0.200}} & \underline{\smash[b]{0.898}} & \underline{\smash[b]{0.878}} & \underline{\smash[b]{0.698}} & \underline{\smash[b]{0.791}} & \underline{\smash[b]{0.716}} & \underline{\smash[b]{0.744}} & \underline{\smash[b]{0.730}} & \underline{\smash[b]{0.858}} & \underline{\smash[b]{0.491}} & \underline{\smash[b]{0.891}} & \underline{\smash[b]{0.964}} \\
 &  & Oracle & \textbf{0.122} & \textbf{0.934} & \textbf{0.925} & \textbf{0.881} & \textbf{0.944} & \textbf{0.893} & \textbf{0.914} & \textbf{0.848} & \textbf{0.922} & \textbf{0.679} & \textbf{1.000} & \textbf{0.998} \\
\addlinespace[1pt]
 & \multirow{3}{*}{7} & Random cases & 0.346 & 0.822 & 0.800 & 0.538 & 0.393 & 0.335 & 0.342 & 0.588 & 0.778 & 0.263 & 0.844 & 0.964 \\
 &  & EndoFinder & \underline{\smash[b]{0.191}} & \underline{\smash[b]{0.883}} & \underline{\smash[b]{0.878}} & \underline{\smash[b]{0.703}} & \underline{\smash[b]{0.807}} & \underline{\smash[b]{0.723}} & \underline{\smash[b]{0.753}} & \underline{\smash[b]{0.731}} & \underline{\smash[b]{0.859}} & \underline{\smash[b]{0.493}} & \underline{\smash[b]{0.932}} & \underline{\smash[b]{0.981}} \\
 &  & Oracle & \textbf{0.142} & \textbf{0.932} & \textbf{0.915} & \textbf{0.856} & \textbf{0.924} & \textbf{0.882} & \textbf{0.900} & \textbf{0.833} & \textbf{0.914} & \textbf{0.657} & \textbf{1.000} & \textbf{1.000} \\
\bottomrule
\end{tabular}%
}
\end{table*}

\begin{table*}[t]
\caption{Single-task and complete-report generation with the proposed method. Check marks indicate the fields requested in each output schema. A dash denotes a metric that is not applicable because the corresponding field was not generated. For \(R\) and \(DPR\), language metrics are computed from the same isolated morphology-description clause; diameter and Paris-type text is excluded from both predictions and references. Bold values indicate the better result between the single-task and complete-report settings for each split and metric.}
\label{tab:task_composition}
\centering
\scriptsize
\setlength{\tabcolsep}{2.0pt}
\renewcommand{\arraystretch}{1.12}
\resizebox{\textwidth}{!}{%
\begin{tabular}{@{}lccc@{\hspace{5pt}}ccc@{\hspace{5pt}}cccc@{\hspace{5pt}}ccc@{}}
\toprule
Split & \(D\) & \(P\) & \(R\) & \multicolumn{3}{c}{Diameter} & \multicolumn{4}{c}{Paris type} & \multicolumn{3}{c}{Morphology description} \\
 & & & & MAE$\downarrow$ & ACC$>$0.5$\uparrow$ & ACC$>$1.0$\uparrow$ & ACC$\uparrow$ & Precision$\uparrow$ & Recall$\uparrow$ & F1$\uparrow$ & BLEU-4$\uparrow$ & ROUGE-L$\uparrow$ & CIDEr$\uparrow$ \\
\midrule
\multirow{4}{*}{Val}
 & \(\checkmark\) &  &  & 0.199 & 0.874 & 0.888 & -- & -- & -- & -- & -- & -- & -- \\
 &  & \(\checkmark\) &  & -- & -- & -- & 0.723 & 0.808 & 0.735 & 0.757 & -- & -- & -- \\
 &  &  & \(\checkmark\) & -- & -- & -- & -- & -- & -- & -- & 0.446 & 0.677 & 0.380 \\
 & \(\checkmark\) & \(\checkmark\) & \(\checkmark\) & \textbf{0.155} & \textbf{0.898} & \textbf{0.917} & \textbf{0.772} & \textbf{0.844} & \textbf{0.756} & \textbf{0.788} & \textbf{0.672} & \textbf{0.826} & \textbf{0.623} \\
\midrule
\multirow{4}{*}{Test}
 & \(\checkmark\) &  &  & 0.225 & \textbf{0.903} & \textbf{0.888} & -- & -- & -- & -- & -- & -- & -- \\
 &  & \(\checkmark\) &  & -- & -- & -- & 0.696 & 0.770 & \textbf{0.724} & 0.740 & -- & -- & -- \\
 &  &  & \(\checkmark\) & -- & -- & -- & -- & -- & -- & -- & 0.425 & 0.673 & 0.347 \\
 & \(\checkmark\) & \(\checkmark\) & \(\checkmark\) & \textbf{0.200} & 0.898 & 0.878 & \textbf{0.698} & \textbf{0.791} & 0.716 & \textbf{0.744} & \textbf{0.629} & \textbf{0.798} & \textbf{0.576} \\
\bottomrule
\end{tabular}
}
\end{table*}

\subsection{Ablation studies}

\noindent\textbf{The two context sources provide complementary gains.}
Figure~\ref{fig:memory_context_effects} summarizes the component ablation across the three reporting tasks, with detailed values provided in Table~\ref{tab:main_results}. In panel A, implicit instruction context and explicit transduction context both improved diameter estimation over Direct, reducing MAE from 0.534 to 0.281 and 0.235, respectively. Combining both contexts further reduced MAE to 0.200 while retaining high agreement at both clinical size thresholds.

Panel B shows a similar but task-dependent pattern for Paris classification. Implicit instruction context increased macro-F1 from 0.135 to 0.513, whereas explicit transduction context reached 0.663. Their combination further increased precision, recall, and macro-F1 to 0.791, 0.716, and 0.744, respectively. The difference between the two individual conditions is consistent with implicit instruction context supplying reusable specialist guidance, whereas explicit transduction context additionally provides query-specific examples.

Panel C shows that both individual contexts also improved report generation. Explicit transduction context produced the stronger single-context result, reaching 0.683 BLEU-4, 0.830 ROUGE-L, and 0.423 CIDEr. Combining both contexts increased these scores to 0.730, 0.858, and 0.491. Together, the two contexts retained the gains of each individual source and produced the best overall result across all three tasks.

\noindent\textbf{Retrieval relevance, depth, and quality jointly determine the value of external context.}

To isolate the effect of retrieval relevance, we replaced the five EndoFinder cases with five randomly sampled training cases while preserving the number and modalities of the additional inputs. On the test set, this intervention increased MM-RAG diameter MAE from 0.235 to 0.400, reduced Paris accuracy from 0.635 to 0.377, and reduced CIDEr from 0.423 to 0.176 (Table~\ref{tab:context_length_control}). Adding Prompt tuning improved the random condition to 0.337 MAE, 0.523 Paris accuracy, and 0.262 CIDEr, but remained below the corresponding relevant-evidence setting. The same ordering was observed on the validation set.

We then varied retrieval depth and evidence quality. Table~\ref{tab:context_quality_topk} compares random, EndoFinder, and oracle evidence at matched values of $K$. Paris hit@$K$ is the proportion of queries with at least one retrieved report containing the target Paris type. Size hit@$K$ analogously measures agreement with the target's binary diameter class at the 0.5~cm threshold. These retrieval metrics quantify whether compatible evidence is available; the downstream metrics measure whether the VLM uses it correctly.

EndoFinder outperformed random retrieval at every matched depth. On the test set, its Paris hit@$K$ increased from 0.637 at $K=1$ to 0.932 at $K=7$, whereas Paris accuracy plateaued near 0.70 from $K=3$ onward. The largest categorical and language gains occurred between one and three EndoFinder cases: macro-F1 increased from 0.699 to 0.758 and CIDEr from 0.428 to 0.489. From $K=3$ to $K=7$, MAE decreased from 0.204 to 0.191, while macro-F1 and CIDEr changed only marginally. Thus, additional cases mainly improved evidence coverage after the first few examples had supplied most of the useful categorical and textual context.

Coverage alone was insufficient. Random retrieval reached a Paris hit rate of 0.844 at $K=7$, but its Paris F1 and CIDEr were only 0.342 and 0.263. Oracle retrieval remained above EndoFinder at every depth; at $K=3$, it achieved a Paris macro-F1 of 0.901 and diameter MAE of 0.130, compared with 0.758 and 0.204 for EndoFinder. This gap identifies remaining headroom in evidence ranking and conflict resolution. The $K=5$ setting achieved the best overall validation performance across diameter estimation, Paris classification, and report generation and was therefore selected as the principal setting.

\noindent\textbf{Joint multi-task reporting improves morphology generation.}
Table~\ref{tab:task_composition} compares the complete diameter--Paris--report (\(DPR\)) schema with the three single-task schemas \(D\), \(P\), and \(R\). Definitions, output rules, and JSON keys for absent fields are removed, while the query image, frozen VLM, top-5 retrieved cases, and decoding settings remain fixed.

Metrics are computed only for requested fields. To compare morphology fairly, the language metrics for \(R\) and \(DPR\) are calculated from the Description clause alone; diameter, Paris type, and their fixed template words are excluded from both predictions and references. The comparison therefore measures how the requested schema changes the shared morphology output rather than rewarding the longer \(DPR\) sentence.

Requesting \(D\) and \(P\) together with \(R\) improved all three morphology metrics on both splits. On the test set, BLEU-4, ROUGE-L, and CIDEr increased from 0.425, 0.673, and 0.347 under \(R\) alone to 0.629, 0.798, and 0.576 under \(DPR\). Because diameter and Paris text is excluded from this scoring, the improvement cannot be attributed to the additional fixed clauses in the complete sentence. Within the same trained model, producing the quantitative and categorical fields therefore provides useful context for morphology generation. The model was trained with complete \(DPR\) reports; the reduced schemas are inference-time interventions on that model rather than independently trained single-task baselines.

\begin{figure*}[p]
\centering
\includegraphics[width=0.95\textwidth]{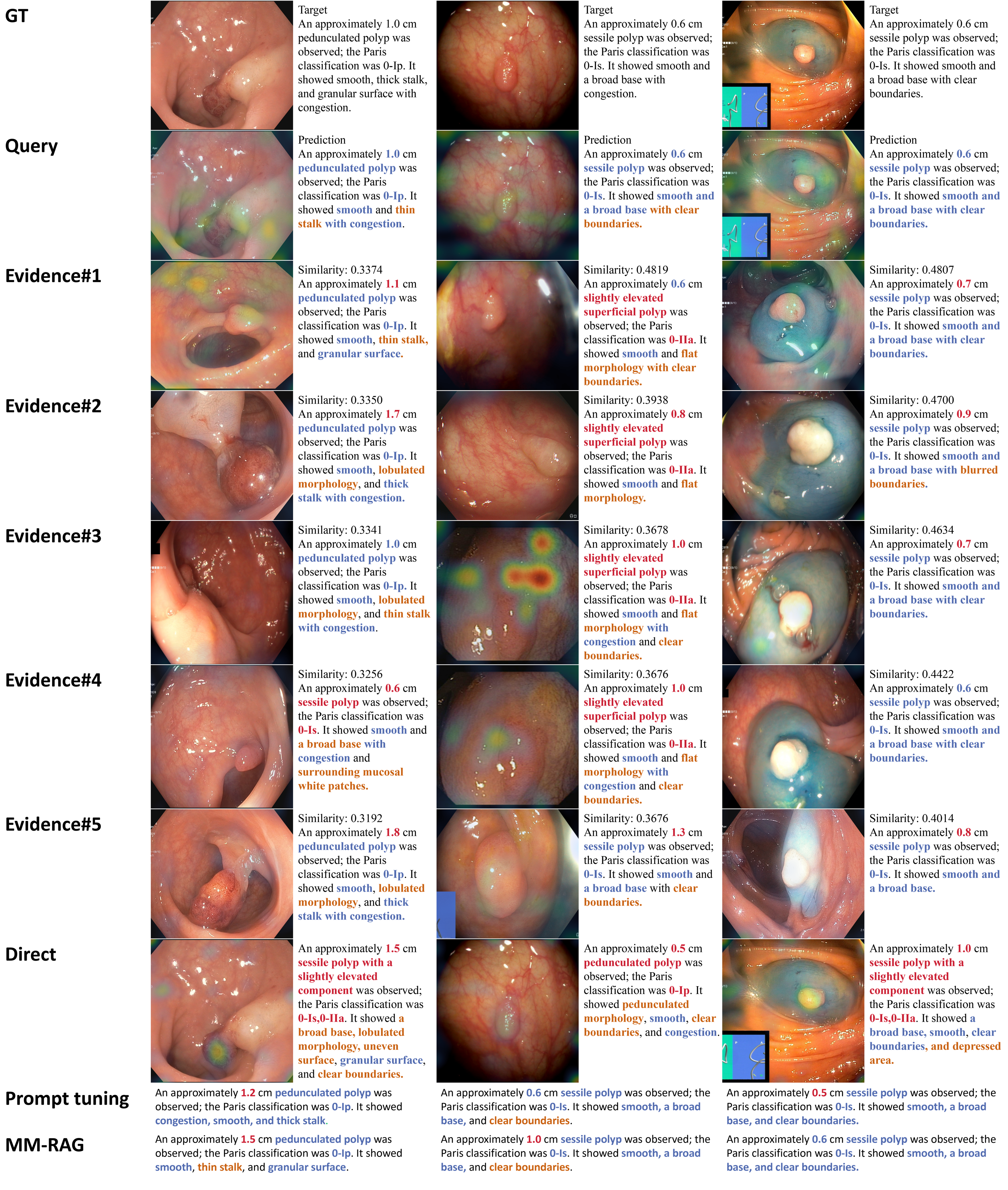}
\caption{Representative context-attribution cases. Each case column contains the query, five retrieved image--report pairs, and predictions from Direct, Prompt tuning, MM-RAG, and Ours. Token-conditioned overlays visualize the attribution of the generated \texttt{Diameter(cm)}, \texttt{ParisType}, and \texttt{Description} values to the query and evidence images; the six image maps within a case share one normalization scale. In the reports, blue denotes content that agrees with the reference, red denotes incorrect content, and orange denotes spurious content not supported by the reference report.}
\label{fig:context_attribution}
\end{figure*}

\subsection{Case studies}

Figure~\ref{fig:context_attribution} links the predictions to the query and retrieved images in three representative cases. The left case contains a pedunculated polyp. Direct inference predicted a composite sessile type, whereas the retrieved set contained four 0-Ip cases and several images with visible stalks. MM-RAG recovered 0-Ip but followed the retrieved thin-stalk descriptions and overestimated diameter. Prompt tuning recovered the target thick-stalk attribute with a smaller size error. Ours combined the correct Paris evidence with the internal context, restoring the target diameter and congestion, although the thin-stalk attribute remained in the final report. The attribution is concentrated on the query image and Evidence~\#1, whose polyp shapes are visually similar. Evidence~\#1 is annotated with a thin stalk, and this attribute appears in both the MM-RAG and combined outputs. The visualization therefore captures both the benefit of a compatible categorical example and the propagation of one conflicting morphology attribute.

In the middle case, the target is a 0.6 cm sessile polyp. The first four retrieved reports describe 0-IIa lesions, whereas only Evidence~\#5 carries the target 0-Is label. Direct inference predicted 0-Ip, and MM-RAG recovered the sessile type but overestimated diameter to 1.0~cm. Prompt tuning recovered the target diameter and Paris type; Ours retained both corrections despite the conflicting retrieved Paris labels. Visual attribution appears mainly in Evidence~\#3--5, with little response in the first two retrieved images, showing that retrieval rank alone did not determine the visual evidence emphasized by the model. In Evidence~\#3 and \#4, attribution is concentrated around the polyp margins; their reports contain smoothness and congestion, which are also present in the target morphology. Evidence~\#5 provides the target 0-Is label and broad-base morphology, and the latter is retained in the final report. These patterns identify the later retrieved cases as the more relevant visual and textual support for this prediction.

The right case provides a complementary supported-retrieval pattern. All five retrieved reports describe sessile polyps, and their diameters and morphology closely match the target. Direct inference instead predicted a larger composite 0-Is+0-IIa lesion with a depressed area. MM-RAG corrected all three fields, and Ours retained the correction. Attribution is strongest on Evidence~\#1, whose image and report closely match the target in size, Paris type, and morphology. These examples do not establish a causal explanation of the model's decision, but they show how compatible and conflicting evidence coexist with the observed corrections.

\begin{figure*}[!t]
\centering
\includegraphics[width=\textwidth]{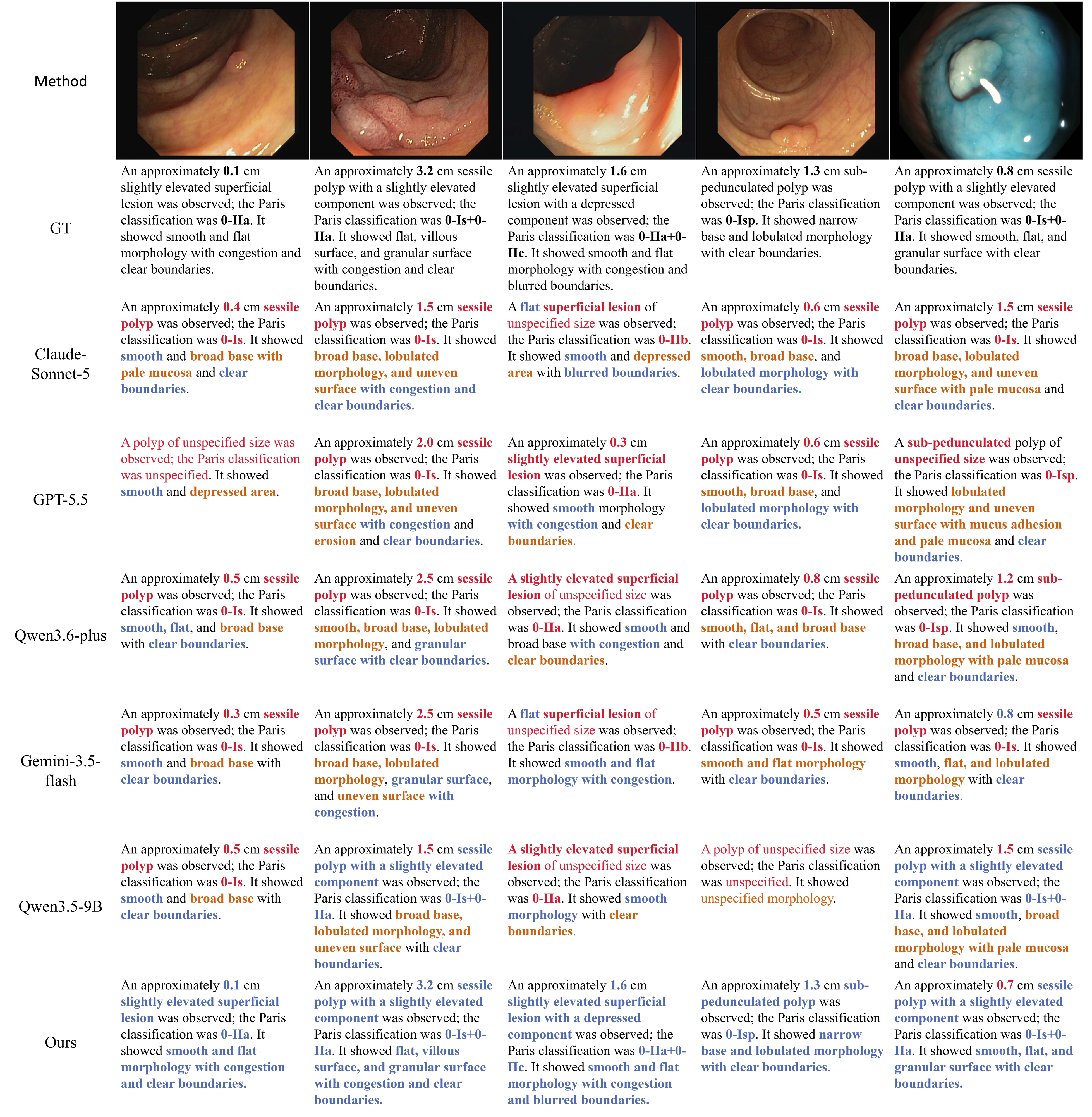}
\caption{Representative case-study comparison on the test set. Each column shows one query case, and the rows compare target reports with direct outputs from general VLMs and the local frozen VLM, followed by the proposed context-fusion setting. Structured JSON predictions are converted into fixed clinical sentences for visual comparison. In model-generated reports, blue denotes content that agrees with the reference, red denotes incorrect content, and orange denotes spurious content not supported by the reference report. In these examples, fluent text can contain clinically important field errors, whereas the proposed setting corrects size or Paris-type content while retaining morphology descriptions.}
\label{fig:general_vlm_case_study}
\end{figure*}

Figure~\ref{fig:general_vlm_case_study} compares the final reports produced by general-purpose VLMs, direct frozen-VLM inference, and the proposed framework. All structured outputs are verbalized by the same fixed rule, so differences arise from predicted content rather than a separate paraphrasing model. Direct models often preserve grammatical structure and common morphology terms while making errors in diameter, Paris type, or a specific descriptive attribute. The combined context setting corrects these fields while retaining coherent report language. The examples also show why ROUGE-L should be read together with diameter and Paris-type metrics: a report may share most words with the reference and still contain a clinically important numerical or categorical error.

\section{Discussion}

This study approaches specialist adaptation as a problem of context organization rather than VLM weight modification. Explicit transduction context supplies query-specific clinical precedents, while implicit instruction context provides a compact condition learned across cases. Because the retrieved cases are supplied as explicit evidence rather than absorbed into model weights, their relevance and report content can be examined for each prediction. This offers a lightweight alternative to specialist systems that train dedicated visual or language components \cite{dentfound2026,conVLM2026,qu2021endoscopyReport,zhang2026samColonPolypGen}.

The task-composition experiment revealed an interaction that is not captured by the main aggregate comparison. Jointly requesting diameter and Paris type improved morphology generation even though their text was excluded from report scoring (Table~\ref{tab:task_composition}). The quantitative and categorical predictions may therefore act as intermediate clinical constraints: estimating lesion scale and gross morphology first can narrow the description that follows. This interpretation is limited to inference-time interventions on a model trained with complete reports, but it suggests that multi-task reporting can contribute more than output convenience.

The hard-case analysis further clarifies when retrieved context is useful. When Finetune failed and the top-ranked retrieved report contained the target Paris type, Ours corrected 70.5\% of those errors while changing 7.0\% of the corresponding Finetune-correct predictions to errors. This imbalance suggests that a relevant precedent can supply missing case-level evidence for difficult categorical decisions without broadly destabilizing predictions that are already correct. The attribution examples also show the boundary of this benefit: compatible evidence can correct size or Paris type, whereas a repeated conflicting attribute can still enter the generated morphology (Fig.~\ref{fig:context_attribution}).

Several limitations define the scope of the present results. Diameter estimation from a monocular image remains uncertain without a calibrated instrument or another scale reference \cite{antonelli2026polypSize}. The structured subset contains 2,056 images, and rare Paris types are under-represented. All adaptation experiments use a single frozen VLM, so the findings do not yet establish cross-backbone generality. The current schema covers diameter, Paris type, and morphology but omits other report fields such as lesion location. Retrieval also relies on one fixed EndoFinder encoder.

Future work should evaluate the framework on larger structured cohorts with calibrated size references, broader Paris-type coverage, and external clinical validation. Applying the same context construction to other frozen VLMs would test whether the observed gains extend across backbones. Field-aware and conflict-aware retrieval could select evidence separately for size, Paris type, morphology, and additional report fields. The oracle results indicate that improved evidence selection could provide further gains without changing the reporting backbone.

\section{Conclusion}

We presented a context-fusion framework that combines query-specific image--report pairs with learned specialist guidance to adapt a frozen VLM for endoscopic polyp reporting. It jointly estimated lesion diameter, classified Paris type, and generated morphology descriptions, achieving the strongest overall balance among the evaluated methods with trainable parameters equal to only 0.006\% of the frozen VLM. Within the evaluated backbone and dataset, these results support context fusion as a lightweight approach to specialist VLM adaptation.

\section*{Data availability}

The annotations used in this study are available through the EndoVL dataset \cite{fu2026endovl} on Figshare at \url{https://doi.org/10.6084/m9.figshare.32320998.v1}. This study uses the 2,056-image polyp-reporting subset defined in Section~\ref{sec:task_dataset}.

\printcredits

\bibliographystyle{unsrtnat}
\bibliography{references}

\end{document}


\begin{CJK*}{UTF8}{gbsn}
\maketitle

\section{Scope of the Supplementary Information}

This document provides additional details on cohort construction, prompt design, multimodal evidence construction, adaptation baselines, and optimization. It also reports validation sensitivity to specialist-token length and controlled analyses of context composition and retrieval-source concentration. Unless stated otherwise, all analyses use the data splits, frozen VLM, EndoFinder case bank, and evaluation pipeline defined in the main manuscript.

\section{Dataset curation and report representation}

\subsection{Release schema and cohort construction}

The source spreadsheet in the public EndoVL release contains 2,709 image-level records and 28 columns. Table~\ref{tab:release_schema} groups the released columns by annotation role. Two columns identify the source collection and image file. The remaining columns describe image type, polyp size and Paris type, optional text, tumour-specific attributes, and 19 morphology findings. The morphology findings are stored as indicator columns rather than as a prewritten report sentence.

\begin{table}[htbp]
\centering
\caption{Column groups in the EndoVL consensus spreadsheet. Column names are reproduced from the released worksheet.}
\label{tab:release_schema}
\small
\begin{tabularx}{\textwidth}{@{}>{\raggedright\arraybackslash}p{0.18\textwidth}>{\raggedright\arraybackslash}p{0.52\textwidth}X@{}}
\toprule
Field group & Released columns & Use in this study \\
\midrule
Image identity & \texttt{lesion\_id/dataset}; \texttt{lesion\_id/Filename} & Locate the image and retain its release source identifier. \\
Polyp report fields & 大小（size\_bin）; 巴黎分型; 额外文字说明 & Size and Paris type define the quantitative and categorical targets; optional free text is not used to construct the target. \\
Tumour fields & 肿瘤-大小; 肿瘤-环周; 肿瘤-JSCCR & Not used because the present task is restricted to polyp-like lesions. \\
Image-level type & 图片类型分类 & Select records labelled 息肉样病变 (polyp-like lesion). \\
Morphology indicators & 伴凹陷; 充血; 光滑; 分叶; 周围黏膜白斑; 平坦; 宽基; 窄基; 粗蒂; 细蒂; 粘液附着; 糜烂; 绒毛状; 脓性分泌物; 苍白; 表面凹凸不平; 边界模糊; 边界清晰; 颗粒状 & Positive indicators are ordered and joined to form the target \texttt{Description}. \\
\bottomrule
\end{tabularx}
\end{table}

The cohort was constructed using two inclusion criteria. First, \texttt{图片类型分类} had to equal \texttt{息肉样病变}. Second, both the released size field and Paris-type field had to be available. Table~\ref{tab:cohort_flow} gives the exact record counts. All 2,059 polyp-like records had a size annotation, whereas three lacked a Paris label. Excluding these records produced the final 2,056-image cohort. A morphology description was generated from the positive indicator columns, and every retained record contained at least one positive finding. Non-polyp and tumour-specific records were excluded.

\begin{table}[htbp]
\centering
\caption{Construction of the structured polyp-reporting cohort from the released consensus spreadsheet.}
\label{tab:cohort_flow}
\begin{tabularx}{\textwidth}{@{}Xrr@{}}
\toprule
Selection stage & Records retained & Records removed at stage \\
\midrule
All released image-level records & 2709 & -- \\
Image type = polyp-like lesion & 2059 & 650 \\
Diameter available & 2059 & 0 \\
Paris type available & 2056 & 3 \\
\midrule
Final structured reporting cohort & 2056 & -- \\
\bottomrule
\end{tabularx}
\end{table}

For each retained row, the released size value was normalized as a centimetre-valued numeric string. Semicolon- or comma-separated Paris labels were normalized to a plus-delimited categorical value, such as \texttt{0-Is+0-IIa}. The 19 morphology columns were traversed in worksheet order; a term was included in \texttt{Description} only when its indicator was positive. This deterministic conversion yielded the three-field target
\begin{equation}
y=\{\texttt{Diameter(cm)},\ \texttt{ParisType},\ \texttt{Description}\}.
\end{equation}

\subsection{Source and label composition of the study cohort}

The 2,056 retained images originate from all nine source identifiers represented in the release. Table~\ref{tab:source_support} reports their allocation after the final split. Each source group occurs in the training, validation, and test sets. The identifiers are retained as recorded in \texttt{lesion\_id/dataset}.

\begin{table}[htbp]
\centering
\caption{Distribution of the filtered cohort by EndoVL release source identifier and data split.}
\label{tab:source_support}
\begin{tabular}{lrrrrr}
\toprule
Source identifier & Training & Validation & Test & Total & Cohort (\%) \\
\midrule
CVC-ClinicDB & 363 & 51 & 99 & 513 & 25.0 \\
images\_C3   & 283 & 41 & 74 & 398 & 19.4 \\
kvasir       & 253 & 37 & 82 & 372 & 18.1 \\
images\_C2   & 157 & 26 & 47 & 230 & 11.2 \\
images\_C1   & 133 & 18 & 39 & 190 & 9.2 \\
images\_C5   & 93  & 12 & 32 & 137 & 6.7 \\
images\_C4   & 69  & 8  & 16 & 93  & 4.5 \\
images\_C6   & 51  & 4  & 8  & 63  & 3.1 \\
CVC-300      & 37  & 9  & 14 & 60  & 2.9 \\
\midrule
Total        & 1439 & 206 & 411 & 2056 & 100.0 \\
\bottomrule
\end{tabular}
\end{table}

The source groups do not provide interchangeable label support. Table~\ref{tab:source_paris_support} shows that only CVC-ClinicDB and kvasir contain all six Paris outcomes. The rare 0-IIa+0-IIc class occurs only in these two sources, whereas CVC-300 contains only 0-Is and 0-IIa cases. Several other sources contain no 0-Ip or no combined Paris class.

\begin{table}[htbp]
\centering
\caption{Paris-type distribution within each EndoVL release source identifier. A dash denotes zero available cases.}
\label{tab:source_paris_support}
\small
\setlength{\tabcolsep}{4.5pt}
\begin{tabular}{lrrrrrrr}
\toprule
Source & 0-Is & 0-Isp & 0-Ip & 0-IIa & \shortstack{0-Is+\\0-IIa} & \shortstack{0-IIa+\\0-IIc} & Total \\
\midrule
CVC-ClinicDB & 141 & 119 & 72 & 145 & 24 & 12 & 513 \\
images\_C3   & 172 & 62  & 80 & 79  & 5  & -- & 398 \\
kvasir       & 157 & 89  & 56 & 51  & 17 & 2  & 372 \\
images\_C2   & 96  & 36  & 33 & 53  & 12 & -- & 230 \\
images\_C1   & 75  & 27  & 9  & 67  & 12 & -- & 190 \\
images\_C5   & 61  & 17  & -- & 53  & 6  & -- & 137 \\
images\_C4   & 36  & 9   & -- & 48  & -- & -- & 93 \\
images\_C6   & 36  & 16  & 10 & 1   & -- & -- & 63 \\
CVC-300      & 7   & --  & -- & 53  & -- & -- & 60 \\
\midrule
Total        & 781 & 375 & 260 & 550 & 76 & 14 & 2056 \\
\bottomrule
\end{tabular}
\end{table}

Diameter coverage is likewise source dependent (Table~\ref{tab:source_diameter_support}). CVC-300 contains no lesion of at least 1.0~cm, and images\_C4 contains none of at least 2.0~cm. The relative contribution of large lesions also differs among the remaining sources. Because no single source covers all Paris types and diameter intervals, all eligible sources were pooled before stratified splitting.

\begin{table}[htbp]
\centering
\caption{Diameter-bin distribution within each EndoVL release source identifier. A dash denotes zero available cases.}
\label{tab:source_diameter_support}
\begin{tabular}{lrrrrr}
\toprule
Source & $<0.5$ & $0.5$--$<1.0$ & $1.0$--$<2.0$ & $\geq2.0$ & Total \\
\midrule
CVC-ClinicDB & 51 & 225 & 203 & 34 & 513 \\
images\_C3   & 73 & 153 & 155 & 17 & 398 \\
kvasir       & 24 & 140 & 165 & 43 & 372 \\
images\_C2   & 5  & 127 & 90  & 8  & 230 \\
images\_C1   & 43 & 71  & 47  & 29 & 190 \\
images\_C5   & 46 & 45  & 13  & 33 & 137 \\
images\_C4   & 37 & 42  & 14  & -- & 93 \\
images\_C6   & 6  & 25  & 30  & 2  & 63 \\
CVC-300      & 49 & 11  & --  & -- & 60 \\
\midrule
Total        & 334 & 839 & 717 & 166 & 2056 \\
\bottomrule
\end{tabular}
\end{table}

The data were divided into 70\% training, 10\% validation, and 20\% test sets. Splitting was stratified by the joint label formed from Paris type and four diameter intervals: $<0.5$, $0.5$--$<1.0$, $1.0$--$<2.0$, and $\geq2.0$~cm. This preserves both Paris-type and coarse lesion-size composition across the three splits. Each image--report pair occurs in only one split. Only the 1,439 training records were used for model optimization and case-bank construction; validation and test images served only as held-out queries, and their reports were not available to the retriever.

Table~\ref{tab:diameter_support} reports the resulting diameter-bin support. Across the full cohort, diameter ranged from 0.1 to 7.0~cm, with a median of 0.8~cm and an interquartile range of 0.5--1.3~cm. The mean was 1.013~cm.

\begin{table}[htbp]
\centering
\caption{Diameter-bin support in the structured reporting subset.}
\label{tab:diameter_support}
\begin{tabular}{lrrrr}
\toprule
Diameter interval (cm) & Training & Validation & Test & Total \\
\midrule
$<0.5$        & 235 & 33 & 66  & 334 \\
$0.5$--$<1.0$ & 587 & 84 & 168 & 839 \\
$1.0$--$<2.0$ & 501 & 72 & 144 & 717 \\
$\geq2.0$     & 116 & 17 & 33  & 166 \\
\midrule
Total         & 1439 & 206 & 411 & 2056 \\
\bottomrule
\end{tabular}
\end{table}

Paris labels were normalized before training and evaluation. A combined label was treated as one complete categorical value rather than as two independent binary labels. Table~\ref{tab:paris_support} reports the exact support of each value. The rare combined classes were retained rather than merged into their constituent classes.

\begin{table}[htbp]
\centering
\caption{Paris-type support in the structured reporting subset.}
\label{tab:paris_support}
\begin{tabular}{lrrrr}
\toprule
Paris type & Training & Validation & Test & Total \\
\midrule
0-Is            & 546 & 79 & 156 & 781 \\
0-Isp           & 263 & 38 & 74  & 375 \\
0-Ip            & 181 & 26 & 53  & 260 \\
0-IIa           & 385 & 55 & 110 & 550 \\
0-Is+0-IIa      & 54  & 7  & 15  & 76  \\
0-IIa+0-IIc     & 10  & 1  & 3   & 14  \\
\midrule
Total           & 1439 & 206 & 411 & 2056 \\
\bottomrule
\end{tabular}
\end{table}

The morphology field is generative, but the instruction supplies the same controlled clinical vocabulary used to construct the references. Multiple findings may be combined in one description, and the model generates the complete field autoregressively. Every retained record has a non-empty description. The median number of positive findings is three (range, one to eight); 1,961 of 2,056 records (95.4\%) contain two to four findings. Table~\ref{tab:morphology_support} gives the frequency of every term. These counts are not mutually exclusive because one report can contain several findings.

\begin{table}[htbp]
\centering
\caption{Frequency of positive morphology findings in the final 2,056-image cohort. Percentages use the number of retained images as the denominator and therefore do not sum to 100\%.}
\label{tab:morphology_support}
\small
\begin{tabular}{lrrlrr}
\toprule
Finding & Count & \% & Finding & Count & \% \\
\midrule
Depressed area & 28 & 1.4 & Congestion & 671 & 32.6 \\
Smooth surface & 1589 & 77.3 & Lobulation & 308 & 15.0 \\
Mucosal white patches & 78 & 3.8 & Flat morphology & 655 & 31.9 \\
Broad base & 781 & 38.0 & Narrow base & 375 & 18.2 \\
Thick stalk & 173 & 8.4 & Thin stalk & 87 & 4.2 \\
Mucus adhesion & 76 & 3.7 & Erosion & 39 & 1.9 \\
Villous surface & 289 & 14.1 & Purulent discharge & 4 & 0.2 \\
Pale mucosa & 9 & 0.4 & Uneven surface & 24 & 1.2 \\
Blurred boundary & 34 & 1.7 & Clear boundary & 781 & 38.0 \\
Granular surface & 289 & 14.1 & & & \\
\bottomrule
\end{tabular}
\end{table}

\section{Task instruction and context construction}

\subsection{Task instruction}

Supplementary Listing S1 reproduces the textual instruction used in the local Qwen3.5-9B experiments. Direct inference and Prompt tuning use the instruction without the \texttt{<RAG>} line. MM-RAG and Ours replace this marker with the evidence blocks defined in Section~\ref{sec:evidence_format}. All conditions otherwise use the same category definitions, output rules, and JSON schema.

\begin{tcolorbox}[enhanced,breakable,title={Supplementary Listing S1. Textual task instruction},
  colback=gray!3,colframe=black!45,boxrule=0.5pt,
  left=6pt,right=6pt,top=5pt,bottom=5pt]
\small\ttfamily
[Medical Knowledge Memory -- Structural Rules]\\[2pt]

【Paris Type（巴黎分型）候选值】\\
- 0-Ip：有蒂隆起（pedunculated）\\
- 0-Isp：部分有蒂/亚有蒂（sub-pedunculated）\\
- 0-Is：无蒂扁平隆起（sessile）\\
- 0-IIa：表浅隆起\\
- 0-IIb：表浅平坦\\
- 0-IIc：表浅凹陷\\[2pt]

【描述提示词（可组合）】\\
伴凹陷、充血、光滑、分叶、周围黏膜白斑、平坦、宽基、窄基、粗蒂、细蒂、粘液附着、糜烂、绒毛状、脓性分泌物、苍白、表面凹凸不平、边界模糊、边界清晰、颗粒状\\[2pt]

【输出规则（必须严格遵守）】\\
1) 只输出一段合法 JSON，不要输出任何解释、推理、标题、markdown、代码块标记（例如 \char96\char96\char96 json）。\\
2) 未知字段用空字符串 \dq{}\dq{}，不要写“无法判断/看不清/N/A/unknown”等。\\
3) Diameter(cm) 只写数字字符串（例如 \dq{}0.6\dq{}），未知则为 \dq{}\dq{}；不要带单位。\\
4) ParisType 允许多个：用英文逗号拼接（例如 \dq{}0-Is,0-IIa\dq{}），未知则为 \dq{}\dq{}。\\[2pt]

\char60 RAG\char62\\[2pt]

请根据图像内容，严格按以下 JSON 结构输出：\\
\{\\
\hspace*{1em}\dq{}Diameter(cm)\dq{}: \dq{}\dq{},\\
\hspace*{1em}\dq{}ParisType\dq{}: \dq{}\dq{},\\
\hspace*{1em}\dq{}Description\dq{}: \dq{}\dq{}\\
\}
\end{tcolorbox}

The \texttt{<RAG>} line is a code-level replacement marker and is not sent to the model. For MM-RAG and Ours, it is replaced by the retrieved cases. For conditions without retrieval, the same instruction is used without the marker. The instruction is presented in Chinese because the source annotations and controlled morphology terms are Chinese; the English terms in parentheses disambiguate the Paris categories.

The hosted general-VLM evaluation used the same no-retrieval user instruction and added the system message shown in Listing~S2. Thus, the hosted models and direct local inference received the same task content, category definitions, and output schema.

\begin{tcolorbox}[enhanced,breakable,title={Supplementary Listing S2. Hosted general-VLM task messages},
  colback=gray!3,colframe=black!45,boxrule=0.5pt,
  left=6pt,right=6pt,top=5pt,bottom=5pt]
\small
\textbf{System message}\par
\ttfamily
You are an expert gastrointestinal endoscopist. Follow the user's output schema strictly and return JSON only.\par
\vspace{4pt}
\rmfamily\textbf{User message}\par
\ttfamily
[Medical Knowledge Memory -- Structural Rules]\\[2pt]

【Paris Type（巴黎分型）候选值】\\
- 0-Ip：有蒂隆起（pedunculated）\\
- 0-Isp：部分有蒂/亚有蒂（sub-pedunculated）\\
- 0-Is：无蒂扁平隆起（sessile）\\
- 0-IIa：表浅隆起\\
- 0-IIb：表浅平坦\\
- 0-IIc：表浅凹陷\\[2pt]

【描述提示词（可组合）】\\
伴凹陷、充血、光滑、分叶、周围黏膜白斑、平坦、宽基、窄基、粗蒂、细蒂、粘液附着、糜烂、绒毛状、脓性分泌物、苍白、表面凹凸不平、边界模糊、边界清晰、颗粒状\\[2pt]

【输出规则（必须严格遵守）】\\
1) 只输出一段合法 JSON，不要输出任何解释、推理、标题、markdown、代码块标记（例如 \char96\char96\char96 json）。\\
2) 未知字段用空字符串 \dq{}\dq{}，不要写“无法判断/看不清/N/A/unknown”等。\\
3) Diameter(cm) 只写数字字符串（例如 \dq{}0.6\dq{}），未知则为 \dq{}\dq{}；不要带单位。\\
4) ParisType 允许多个：用英文逗号拼接（例如 \dq{}0-Is,0-IIa\dq{}），未知则为 \dq{}\dq{}。\\[2pt]

请根据图像内容，严格按以下 JSON 结构输出：\\
\{\\
\hspace*{1em}\dq{}Diameter(cm)\dq{}: \dq{}\dq{},\\
\hspace*{1em}\dq{}ParisType\dq{}: \dq{}\dq{},\\
\hspace*{1em}\dq{}Description\dq{}: \dq{}\dq{}\\
\}
\end{tcolorbox}

\subsection{Explicit transduction context: format and selection}
\label{sec:evidence_format}

Each retrieved image remains linked to its structured report and EndoFinder cosine-similarity score. A human-readable version of one explicit transduction context item is

\begin{tcolorbox}[title={Supplementary Listing S3. One retrieved case},
  colback=gray!3,colframe=black!45,boxrule=0.5pt,left=6pt,right=6pt,top=5pt,bottom=5pt]
\small\ttfamily
[Similar Case]\\
- Similarity/Relevance: $s_i$\\
- Evidence image: Retrieved evidence image $i$\\
- Description: \{\dq{}Diameter(cm)\dq{}: \dq{}$d_i$\dq{}, \dq{}ParisType\dq{}: \dq{}$p_i$\dq{},\\
\hspace*{8.8em}\dq{}Description\dq{}: \dq{}$r_i$\dq{}\}
\end{tcolorbox}

At inference, five cases are ordered by decreasing EndoFinder cosine similarity, and their images and reports are supplied together. The input order is: target-image tokens, learned specialist tokens when enabled, task rules, retrieved image--report evidence, and the output schema. The specialist tokens parameterize the implicit instruction context. They have no discrete textual form and are therefore not shown in Listing~S1.

When training Ours, the ten nearest training cases are first identified for each query, and five are sampled from this set. At validation and test time, the five highest-ranked cases are used. A training query is excluded from retrieving itself from the case bank.

\section{Adaptation baselines and implementation details}

\subsection{Trainable locations in Qwen3.5-9B}

Figure~\ref{fig:qwen_adaptation_locations} summarizes the visual-to-language path used in the Qwen3.5-9B experiments. The vision encoder has 27 transformer blocks with a hidden width of 1,152. Image patches are spatially merged in $2\times2$ groups. The resulting 4,608-dimensional concatenated features pass through the visual-language merger and are projected to the 4,096-dimensional language-model space. The language model contains 32 decoder layers.

\begin{figure}[H]
\centering
\begin{tikzpicture}[
  font=\small,
  node distance=4.5mm,
  every node/.style={inner sep=2.5pt},
  flow/.style={-{Latex[length=2.2mm]},line width=0.75pt,draw=black!65},
  frozen/.style={draw=blue!65!black,fill=blue!7,rounded corners=2pt,minimum height=19mm,align=center,font=\scriptsize},
  tuned/.style={draw=orange!85!black,fill=orange!13,rounded corners=2pt,minimum height=19mm,align=center,font=\scriptsize},
  plain/.style={draw=black!55,fill=white,rounded corners=2pt,minimum height=19mm,align=center,font=\scriptsize},
  context/.style={draw=black!50,fill=black!2,rounded corners=2pt,minimum height=12mm,align=center,font=\scriptsize}
]
\node[plain,text width=1.75cm] (images) {Query and retrieved\\images};
\node[frozen,right=of images,text width=2.25cm] (vision) {Vision encoder\\27 blocks, $h=1152$\\$16\times16$ patches};
\node[tuned,right=of vision,text width=2.35cm] (merger) {Visual-language\\merger\\$2\times2$ spatial\\merge\\$4608\!\rightarrow\!4096$};
\node[frozen,right=of merger,text width=2.65cm] (decoder) {Qwen3.5 language\\model\\32 layers, $h=4096$\\[1mm]\colorbox{violet!12}{\textcolor{violet!75!black}{\scriptsize LoRA adapters}}};
\node[plain,right=of decoder,text width=1.45cm] (report) {Structured\\report};

\draw[flow] (images) -- (vision);
\draw[flow] (vision) -- (merger);
\draw[flow] (merger) -- (decoder);
\draw[flow] (decoder) -- (report);

\node[context,below=8mm of decoder,text width=3.85cm] (text) {Task instruction, retrieved reports,\\output schema, and specialist tokens};
\draw[flow] (text.north) -- (decoder.south);

\node[below=7mm of text,anchor=center,font=\scriptsize] (legend) {\textcolor{blue!65!black}{\rule{8pt}{8pt}} frozen base\hspace{13pt}\textcolor{orange!85!black}{\rule{8pt}{8pt}} trained for Finetune\hspace{13pt}\textcolor{violet!75!black}{\rule{8pt}{8pt}} LoRA adapters};
\end{tikzpicture}
\caption{Locations of the two weight-adaptation baselines in Qwen3.5-9B. Finetune updates only the visual-language merger that maps merged visual features into the language-model embedding space. LoRA leaves the original weights fixed and inserts low-rank adapters into language-attention projection layers. Repeated transformer and decoder blocks are condensed for clarity.}
\label{fig:qwen_adaptation_locations}
\end{figure}

For the Finetune baseline, all parameters are frozen except \texttt{model.visual.merger.norm}, \texttt{linear\_fc1}, and \texttt{linear\_fc2}. Their weight and bias tensors contain 40.119~M trainable parameters. The vision transformer and language model are unchanged, confining adaptation to the interface between visual features and language tokens.

For the LoRA baseline, the original Qwen parameters are also frozen. Rank-8 adapters are inserted into the language model's full-attention projections (\texttt{q\_proj}, \texttt{k\_proj}, \texttt{v\_proj}, and \texttt{o\_proj}) and linear-attention projections (\texttt{in\_proj\_qkv}, \texttt{in\_proj\_z}, \texttt{in\_proj\_b}, \texttt{in\_proj\_a}, and \texttt{out\_proj}). We use $\alpha=16$, dropout 0.05, and no trainable bias, yielding 9.056~M trainable parameters. Table~\ref{tab:adaptation_details} contrasts these baselines with the context-based conditions.

\begin{table}[H]
\centering
\caption{Trainable state and adaptation location for the Qwen3.5-9B conditions.}
\label{tab:adaptation_details}
\small
\begin{tabularx}{\textwidth}{@{}l>{\raggedright\arraybackslash}p{0.30\textwidth}>{\raggedright\arraybackslash}Xr@{}}
\toprule
Condition & Trainable location & Setting & Parameters (M) \\
\midrule
Direct / MM-RAG & None & Frozen Qwen3.5-9B; MM-RAG changes only inference context & 0 \\
Prompt tuning / Ours & Specialist-token embeddings & 128 continuous embeddings of width 4,096 & 0.524 \\
LoRA & Language-attention projections & Rank 8; $\alpha=16$; dropout 0.05; bias disabled & 9.056 \\
Finetune & Visual-language merger & Norm and two linear layers from merged visual features to $h=4096$ & 40.119 \\
\bottomrule
\end{tabularx}
\end{table}

\subsection{Framework optimization and decoding}

Table~\ref{tab:implementation} lists the settings used for the proposed framework. Its implicit instruction context is parameterized by one trainable specialist-token embedding matrix.

\begin{table}[htbp]
\centering
\small
\renewcommand{\arraystretch}{0.88}
\caption{Implementation settings for the proposed framework.}
\label{tab:implementation}
\begin{tabularx}{\textwidth}{@{}>{\raggedright\arraybackslash}p{0.35\textwidth}X@{}}
\toprule
Setting & Value \\
\midrule
Frozen generator & Qwen3.5-9B \\
Image processing & Native input resolution; Qwen variable-resolution processor \\
Retriever & Frozen EndoFinder ViT; 1,024-dimensional L2-normalized CLS features \\
Default retrieved cases & 5 image-report pairs \\
Training retrieval & 5 cases sampled from the 10 nearest training cases \\
Specialist-token length & 128 tokens \\
Hidden dimension & 4,096 \\
Trainable parameters & 524,288 \\
Initialization & Gaussian, standard deviation 0.02 \\
Optimizer & AdamW \\
Epochs & 20 \\
Peak learning rate & $1\times10^{-4}$ \\
Weight decay & 0.1 \\
Learning-rate schedule & OneCycle; warm-up fraction 0.1 \\
Micro-batch / gradient accumulation & 1 / 16 (effective batch size 16) \\
Gradient clipping & 1.0 \\
Numerical precision & bfloat16 mixed precision \\
Gradient checkpointing & Enabled \\
Generation & Greedy decoding; one beam; maximum 128 new tokens \\
Validation/test batch size & 1 \\
Hardware & One NVIDIA A100 80GB GPU \\
\bottomrule
\end{tabularx}
\end{table}

Only the specialist-token embedding matrix is updated for Prompt tuning and Ours. The Qwen parameters, visual processor, EndoFinder encoder, and case-bank embeddings remain fixed. The checkpoint with the lowest validation loss is used for evaluation. Direct inference and MM-RAG require no optimization.

\FloatBarrier
\subsection{Report verbalization for language evaluation}

Predictions and references are first normalized to the same three-field JSON schema and then converted to English with one deterministic verbalizer. Supplementary Listing S4 gives the sentence form used before BLEU-4, ROUGE-L, and CIDEr are calculated.

\begin{tcolorbox}[title={Supplementary Listing S4. Deterministic English report form},
  colback=gray!3,colframe=black!45,boxrule=0.5pt,left=6pt,right=6pt,top=5pt,bottom=5pt]
\small\ttfamily
An approximately $\langle$Diameter(cm)$\rangle$ cm $\langle$Paris-derived lesion phrase$\rangle$ was observed;\\
the Paris classification was $\langle$ParisType$\rangle$.\\
It showed $\langle$grouped Description phrase$\rangle$.
\end{tcolorbox}

The Paris-derived lesion phrase is fixed by Table~\ref{tab:verbalizer_paris}. Description terms are mapped to their canonical English forms and grouped as appearance, accompanying findings, and boundary findings. Terms retain the worksheet order within each group and are joined deterministically. Missing diameter or Paris values are rendered as ``unspecified.'' The same rule is applied to predictions and references; no language model is used during verbalization. BLEU-4, ROUGE-L, and CIDEr are computed with the implementations in \texttt{MedCaption/evalcap}; CIDEr is divided by 10 for display on the scale used in the main manuscript.

\begin{table}[htbp]
\centering
\caption{Paris-type phrases used by the deterministic English verbalizer.}
\label{tab:verbalizer_paris}
\small
\begin{tabular}{ll}
\toprule
Paris type & Lesion phrase \\
\midrule
0-Ip & pedunculated polyp \\
0-Isp & sub-pedunculated polyp \\
0-Is & sessile polyp \\
0-IIa & slightly elevated superficial polyp \\
0-IIb & flat superficial polyp \\
0-IIc & slightly depressed superficial polyp \\
0-Is+0-IIa & sessile polyp with a slightly elevated component \\
0-IIa+0-IIc & slightly elevated polyp with a depressed component \\
\bottomrule
\end{tabular}
\end{table}

\section{Sensitivity to specialist-token length}

We varied the number of learned continuous tokens while retaining the same frozen Qwen3.5-9B backbone, training data, task instruction, and optimization schedule. This capacity study uses Prompt tuning without retrieved cases and therefore isolates the capacity of the implicit instruction context. Only validation results are reported because this experiment was used to select the token length before final test evaluation.

Table~\ref{tab:prompt_length} uses the same ten metrics as the main result table. The parameter count is $M\times4{,}096$, where $M$ is the token count. Bold and underlined values denote the best and second-best distinct validation results within each metric. The dagger identifies the length used in the main experiments.

\begin{table}[H]
\centering
\caption{Validation sensitivity to the number of specialist tokens used for Prompt tuning. All values are computed from the stored per-case JSON predictions with the same normalization, verbalization, and metric implementation used for the main results.}
\label{tab:prompt_length}
\footnotesize
\renewcommand{\arraystretch}{0.86}

\textbf{A. Diameter estimation}\\[3pt]
\begin{tabular}{rrrrr}
\toprule
Tokens & Params (M) & MAE$\downarrow$ & ACC$>$0.5$\uparrow$ & ACC$>$1.0$\uparrow$ \\
\midrule
8   & 0.033 & 0.324 & 0.811 & 0.811 \\
16  & 0.066 & 0.335 & 0.854 & 0.767 \\
32  & 0.131 & 0.287 & 0.859 & 0.835 \\
64  & 0.262 & 0.257 & \second{0.883} & \second{0.859} \\
96  & 0.393 & \second{0.245} & \best{0.888} & \best{0.869} \\
128$^{\dagger}$ & 0.524 & \best{0.244} & 0.864 & \best{0.869} \\
160 & 0.655 & 0.261 & 0.879 & 0.840 \\
256 & 1.049 & 0.254 & \second{0.883} & 0.845 \\
\bottomrule
\end{tabular}

\vspace{4pt}
\textbf{B. Paris-type classification}\\[3pt]
\begin{tabular}{rrrrr}
\toprule
Tokens & ACC$\uparrow$ & Precision$\uparrow$ & Recall$\uparrow$ & F1$\uparrow$ \\
\midrule
8   & 0.578 & 0.460 & 0.395 & 0.407 \\
16  & 0.558 & 0.327 & 0.325 & 0.322 \\
32  & 0.626 & \second{0.571} & 0.422 & 0.440 \\
64  & \best{0.660} & 0.570 & \best{0.507} & \best{0.530} \\
96  & 0.626 & 0.506 & 0.472 & 0.485 \\
128$^{\dagger}$ & 0.641 & 0.505 & 0.463 & 0.476 \\
160 & 0.626 & 0.499 & 0.455 & 0.471 \\
256 & \second{0.646} & \best{0.607} & \second{0.491} & \second{0.527} \\
\bottomrule
\end{tabular}

\vspace{4pt}
\textbf{C. Complete-report generation}\\[3pt]
\begin{tabular}{rrrr}
\toprule
Tokens & BLEU-4$\uparrow$ & ROUGE-L$\uparrow$ & CIDEr$\uparrow$ \\
\midrule
8   & 0.614 & 0.799 & 0.284 \\
16  & 0.623 & 0.803 & 0.317 \\
32  & 0.648 & 0.816 & 0.343 \\
64  & \best{0.671} & \best{0.831} & 0.380 \\
96  & 0.663 & 0.826 & 0.373 \\
128$^{\dagger}$ & 0.668 & 0.826 & \best{0.387} \\
160 & 0.661 & 0.825 & 0.378 \\
256 & \second{0.669} & \second{0.829} & \second{0.381} \\
\bottomrule
\end{tabular}
\end{table}

Performance improved substantially between 16 and 64 tokens and then plateaued, with different capacities favoring different metrics. Sixty-four tokens gave the highest Paris accuracy, recall, F1, BLEU-4, and ROUGE-L. The 128-token setting gave the lowest diameter MAE and the highest CIDEr while retaining competitive threshold and classification results. We therefore fixed the specialist-token length at 128 for the subsequent multimodal-context experiments.

\section{Controlled context analyses}

\subsection{Context variants}

Table~\ref{tab:context_variants} specifies the information sources present in each controlled condition. All conditions retain the target image, task rules, output schema, frozen Qwen backbone, and decoding settings.

\begin{table}[H]
\centering
\caption{Information sources used by the principal controlled variants.}
\label{tab:context_variants}
\footnotesize
\setlength{\tabcolsep}{3.5pt}
\begin{tabularx}{\textwidth}{@{}>{\raggedright\arraybackslash}p{0.27\textwidth}ccc>{\raggedright\arraybackslash}X@{}}
\toprule
Condition & \shortstack{Specialist\\tokens} & \shortstack{Retrieved\\images} & \shortstack{Retrieved\\reports} & Evidence selection \\
\midrule
Direct frozen VLM & -- & -- & -- & None \\
Prompt tuning & \yes & -- & -- & None \\
MM-RAG & -- & \yes & \yes & EndoFinder top-5 \\
Ours & \yes & \yes & \yes & EndoFinder top-5 \\
Random MM-RAG & -- & \yes & \yes & Five random training cases \\
Prompt tuning + random MM-RAG & \yes & \yes & \yes & Five random training cases \\
Quality/depth control & \yes & \yes & \yes & Random, EndoFinder, or label oracle at matched $K$ \\
\bottomrule
\end{tabularx}
\end{table}

For the random-context input-length control, random cases use the same image--report format as EndoFinder cases. The same random assignment is retained across paired conditions. For retrieval-depth experiments, $K\in\{1,3,5,7\}$ and each evidence source supplies exactly $K$ cases.

The label oracle is an analytical upper-bound condition. Its ranking score for training case $i$ and query $q$ is
\begin{equation}
S(q,i)=0.40\,\mathbb{1}[p_q=p_i]
+0.35\max\!\left(0,1-\frac{|d_q-d_i|}{2}\right)
+0.25\,J(r_q,r_i),
\end{equation}
where $J(\cdot,\cdot)$ is Jaccard overlap between normalized morphology-term sets. The oracle quantifies the headroom available if more label-compatible evidence could be ranked near the front of the context.

\subsection{Source composition of retrieved evidence}

Source concentration was calculated from the Top-5 evidence associated with each validation and test query. A source is the release identifier given by the parent directory of the image path (Table~\ref{tab:source_support}). Random denotes the matched context control containing five randomly selected training cases for each query.

Table~\ref{tab:retrieval_source_analysis}A reports five complementary measures. ``Same-source cases'' is the percentage of all evidence cases that share the query source. ``Top-1 same'' is the percentage of queries whose first evidence case has the same source. ``Any of five'' and ``all five'' indicate whether at least one or all five evidence cases share the query source. ``Mean sources'' is the average number of distinct source identifiers among the five cases.

Panels B and C examine whether unequal label distributions could explain the source concentration. Panel B stratifies queries by their ground-truth Paris type. Panel C compares each query's ground-truth structured fields with the report fields of every evidence case. A Paris match requires exact equality between the two normalized Paris types; a combined label is treated as one complete category. A diameter-bin match requires both diameters to fall in the same interval: $<0.5$, $0.5$--$<1.0$, $1.0$--$<2.0$, or $\geq2.0$~cm. ``Subset share'' is the percentage of all evidence cases that meet the row condition. The adjacent same-source percentage is calculated only within that subset.

\begin{table}[H]
\centering
\caption{Source composition of Top-5 evidence. Panel A summarizes all validation and test queries. Panel B stratifies test queries by Paris type. Panel C conditions the test-set analysis on evidence--query label agreement. $\Delta$ denotes EndoFinder minus Random in percentage points. The 0-IIa+0-IIc test stratum contains three queries.}
\label{tab:retrieval_source_analysis}
\scriptsize
\setlength{\tabcolsep}{3.8pt}
\textbf{A. Overall source composition}\par\smallskip
\begin{tabular}{llrrrrr}
\toprule
Split & Evidence & \shortstack{Same-source\\cases (\%)} & \shortstack{Top-1\\same (\%)} & \shortstack{Any of\\five (\%)} & \shortstack{All five\\(\%)} & \shortstack{Mean\\sources} \\
\midrule
Validation & EndoFinder & 63.0 & 75.2 & 93.2 & 32.0 & 2.25 \\
           & Random     & 15.3 & 17.0 & 55.3 & 0.0  & 3.72 \\
Test       & EndoFinder & 62.1 & 71.3 & 92.9 & 32.6 & 2.25 \\
           & Random     & 15.5 & 14.1 & 52.8 & 0.2  & 3.61 \\
\bottomrule
\end{tabular}

\vspace{6pt}
\textbf{B. Test queries stratified by Paris type}\par\smallskip
\begin{tabular}{lrrrrrr}
\toprule
Paris type & $n$ & \shortstack{EndoFinder\\same-source (\%)} & \shortstack{Random\\same-source (\%)} & $\Delta$ & \shortstack{EndoFinder\\mean sources} & \shortstack{Random\\mean sources} \\
\midrule
0-Is        & 156 & 57.8 & 15.3 & 42.6 & 2.35 & 3.69 \\
0-Isp       & 74  & 58.6 & 14.6 & 44.1 & 2.36 & 3.54 \\
0-Ip        & 53  & 56.2 & 16.2 & 40.0 & 2.53 & 3.49 \\
0-IIa       & 110 & 70.7 & 15.8 & 54.9 & 1.95 & 3.63 \\
0-Is+0-IIa  & 15  & 73.3 & 17.3 & 56.0 & 2.13 & 3.47 \\
0-IIa+0-IIc & 3   & 100.0 & 20.0 & 80.0 & 1.00 & 3.00 \\
\bottomrule
\end{tabular}

\vspace{6pt}
\textbf{C. Test evidence conditioned on label agreement}\par\smallskip
\begin{tabular}{lrrrr}
\toprule
Evidence subset & \shortstack{EndoFinder\\subset share (\%)} & \shortstack{EndoFinder\\same-source (\%)} & \shortstack{Random\\subset share (\%)} & \shortstack{Random\\same-source (\%)} \\
\midrule
Paris match          & 55.1 & 76.7 & 28.2 & 15.9 \\
Paris mismatch       & 44.9 & 44.1 & 71.8 & 15.4 \\
Paris and diameter-bin match & 41.9 & 83.3 & 12.2 & 17.1 \\
\bottomrule
\end{tabular}
\end{table}

EndoFinder retrieves more same-source evidence than Random on both splits (Table~\ref{tab:retrieval_source_analysis}A), although its test-set Top-5 still spans 2.25 sources on average. The difference persists within every Paris type (Panel B) and among cases matched by Paris type and diameter interval (Panel C). Thus, label imbalance alone does not account for the observed source concentration. The retrieval representation appears to retain source-associated appearance together with lesion similarity.

\end{CJK*}